\documentclass[journal]{IEEEtai}

\usepackage[colorlinks,urlcolor=blue,linkcolor=blue,citecolor=blue]{hyperref}
\let\labelindent\relax
\usepackage{enumitem}
\usepackage{color,array}
\usepackage{multicol}
\usepackage{times}
\usepackage{epsfig}
\usepackage{graphicx}
\usepackage{amssymb}
\usepackage{amsmath}

\usepackage{multirow}
\usepackage{tabularx}
\usepackage{booktabs}
\usepackage{amsfonts}
\usepackage{nicefrac}
\usepackage{microtype}

\usepackage{hyperref}
\usepackage{xcolor}
\usepackage{kotex}
\usepackage{wrapfig}
\usepackage{algorithm, algpseudocode}%
\usepackage{rotating}
\usepackage{graphicx}

\begin{document}

\title{Information-Theoretic Visual Explanation for Black-Box Classifiers}

\author{Jihun~Yi,
        Eunji~Kim,
        Siwon~Kim, 
        and~Sungroh~Yoon,~\IEEEmembership{Senior Member,~IEEE}
\thanks{J. Yi, E. Kim, S. Kim, and S. Yoon are with Seoul National University, South Korea.}
\thanks{S. Yoon is also with ASRI, INMC, ISRC, and Institute of Engineering Research, Seoul National University, Seoul 08826, Korea.}
\thanks{Correspondence should be addressed to S. Yoon (sryoon@snu.ac.kr).}
}

\maketitle

\newcommand{\Xvec}{\mathbf{X}}
\newcommand{\xvec}{\mathbf{x}}
\newcommand{\avec}{\mathbf{a}}
\newcommand{\bvec}{\mathbf{b}}
\newcommand{\mvec}{\mathbf{m}}

\newcommand{\xr}{\xvec_r}
\newcommand{\xmr}{\xvec_{\setminus r}}
\newcommand{\Xr}{\Xvec_r}

\newcommand{\xii}{x_i}
\newcommand{\xmi}{\xvec_{\setminus i}}
\newcommand{\Xii}{X_i}
\newcommand{\xref}{x_{\rm{ref}}}

\newcommand{\xtmr}{\tilde{\xvec}_{\setminus r}}
\newcommand{\xtmi}{\tilde{\xvec}_{\setminus i}}
\newcommand{\xhmi}{\widehat{\xvec}_{\setminus i}}
\newcommand{\xhmr}{\widehat{\xvec}_{\setminus r}}

\newcommand{\xhr}{\widehat{\xvec}_r}
\newcommand{\xhi}{\widehat{x}_i}
\newcommand{\xtr}{\tilde{\xvec}_r}
\newcommand{\xti}{\tilde{x}_i}

\newcommand{\etal}{\textit{et al.}}
\newcommand{\ie}{\textit{i}.\textit{e}.}
\newcommand{\eg}{\textit{e}.\textit{g}.}

\newcommand{\textblue}{\textcolor{blue}}
\newcommand{\textred}{\textcolor{red}}

\newcommand{\pmimap}{\mathbf{m}_{\rm{pmi}}}
\newcommand{\igmap}{\mathbf{m}_{\rm{ig}}}

\newcommand{\kl}{\textup{D}_\textup{KL}}
\newcommand{\defeq}{\overset{\underset{\mathrm{def}}{}}{=}}
\newcommand{\cond}{\,|\,}

\newcommand\numberthis{\addtocounter{equation}{1}\tag{\theequation}}

\newcommand{\addcite}{\textred{(add cite)}~}


\begin{abstract}
In this work, we attempt to explain the prediction of any black-box classifier from an information-theoretic perspective.
For each input feature, we compare the classifier outputs with and without that feature using two information-theoretic metrics.
Accordingly, we obtain two attribution maps---an information gain \textbf{(IG) map} and a point-wise mutual information \textbf{(PMI) map}.
IG map provides a class-\textit{independent} answer to ``How informative is each pixel?", and PMI map offers a class-\textit{specific} explanation of ``How much does each pixel support a specific class?" 
Compared to existing methods, our method improves the correctness of the attribution maps in terms of a quantitative metric.
We also provide a detailed analysis of an ImageNet classifier using the proposed method, and the code is available online\footnote{\url{https://github.com/nuclearboy95/XAI-Information-Theoretic-Explanation}}.

\end{abstract}

\begin{IEEEImpStatement}
Considering an explosive number of deployments of neural networks in real-world application, understanding their mechanism is an important issue.
This work presents a novel method for interpreting a prediction of an image classifier.
The proposed method provides more correct interpretation compared to existing methods, and it offers unique explanation not available in existing methods.
We believe that the adoption of the proposed method will benefit the researchers, users, and society in that the method enables a deeper understanding of neural networks.
\end{IEEEImpStatement}


\section{Introduction}
\IEEEPARstart{I}{mage} classifiers using deep neural networks demonstrate a high level of performance \cite{hu2018squeeze,simonyan2014very}, but they lack human interpretability; they provide accurate classification results but the reasoning behind them is not accessible.
This black-box property hinders their prediction from convincing a human user.
Therefore, making a neural network interpretable is a prerequisite for high-stakes applications.
A popular approach for explaining the prediction of an image classifier is to generate an \textit{attribution} map \cite{shrikumar2017learning,sundararajan2017axiomatic}, which is the quantified contribution of each input pixel to the prediction.
Overlapping the map with an input image offers an intuitive and easily interpretable visualization that highlights a salient region of the image.

To estimate an input feature's importance to the classifier output, it is reasonable to compare the outputs \textit{with} and \textit{without} that feature.
If removing feature $\xii$ from input $\xvec$ results in a significant change in the classifier output, it can be concluded that $\xii$ plays an important role in the classifier prediction $p_{\theta}(Y \cond \xvec)$ for a label $Y$.
A group of recent explanation methods called \textit{perturbation-based methods}~ \cite{zeiler2014visualizing,zintgraf2017visualizing} takes this approach.
First, the methods calculate a prediction for a partially removed input ($p_{\theta}(Y \cond \xmi)$) and compare it with the original prediction $p_{\theta}(Y \cond \xvec)$ to quantify the contribution of $\xii$.
More formally, a perturbation-based method consists of the following two steps:
\begin{enumerate}[leftmargin=1.5cm, label=Step \arabic*:]
    \item Calculate $p_{\theta}(Y \cond \xmi)$
    \item Calculate $m_i=\textrm{Diff} \left (p_{\theta}(Y \cond \xvec), p_{\theta}(Y \cond \xmi) \right )$,
\end{enumerate}
where the $\textrm{Diff}(\cdot)$ function compares predictions with and without feature $\xii$, and the specific function varies among methods.
The quantified difference between the two predictions becomes $m_i$, which is a contribution of $\xii$.

\begin{figure}[t]
    \centering
    \includegraphics[width=\linewidth]{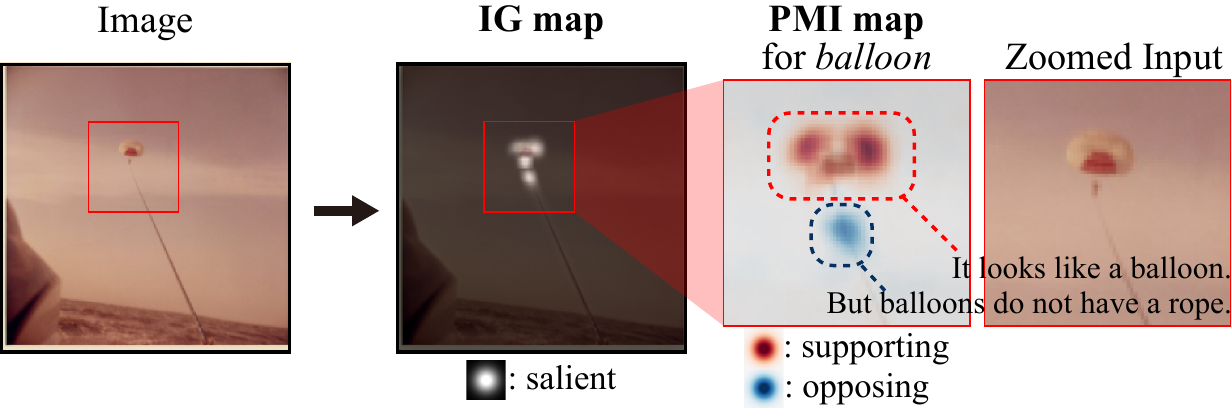}
    \vspace{-5pt}
    \caption{\textbf{The proposed IG and PMI maps.}
    VGG19~\cite{simonyan2014very} classified the parachute image as a balloon.
    The \textbf{IG map} highlights salient regions (bright) for the classifier, and the \textbf{PMI map} indicates evidence for (red) and against (blue) the \textit{balloon} class from the viewpoint of the classifier.
    The orange fabric of the parachute is identified as supporting evidence for \textit{balloon} class because it appears like a part of a balloon.
    However, the rope is highlighted as opposing evidence because balloons do not have a rope.
    }
    \label{fig:example}
    \vspace{-10pt}
\end{figure}

\begin{figure*}[t]
    \centering
    \includegraphics[width=0.9\textwidth]{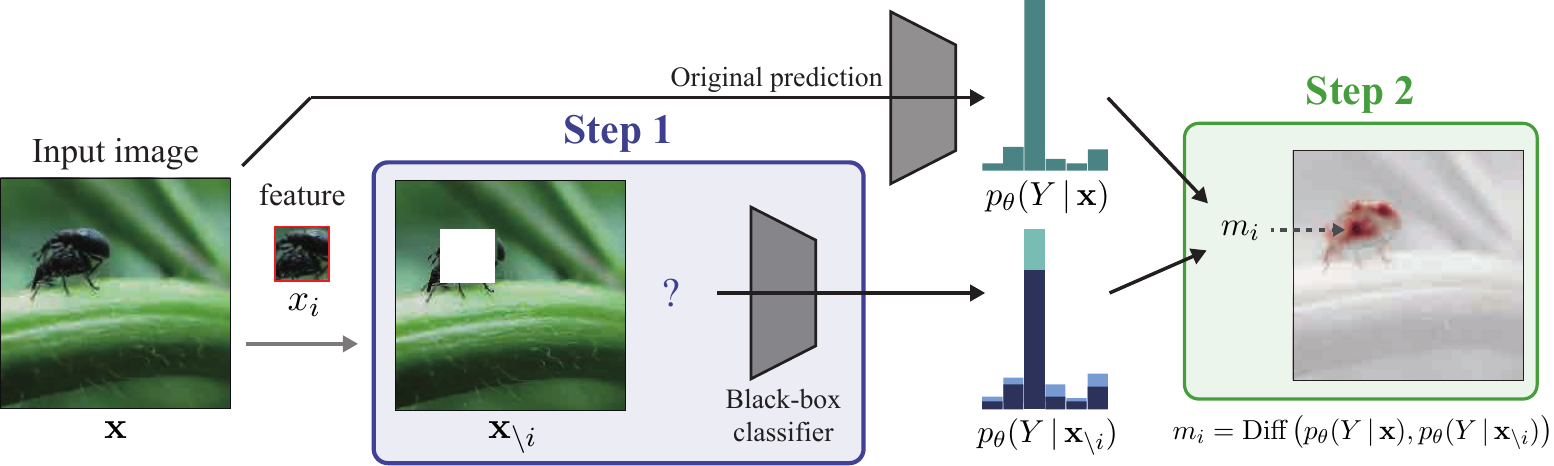}
    \vspace{-5pt}
    \caption{\textbf{The overall flow of perturbation-based methods.}
    For each input area $\xii$, perturbation-based methods estimate the classifier output without $\xii$, $p_{\theta}(Y \cond \xmi)$. In next step, they compare the estimated $p_{\theta}(Y \cond \xvec)$ with the original prediction $p_{\theta}(Y \cond \xvec)$ and quantify the attribution of $\xii$.
    }
    \label{fig:perturbation_methods}
\end{figure*}

The existing perturbation-based methods~\cite{zeiler2014visualizing,zintgraf2017visualizing} proposed approaches for each step, but their approaches for Step 1 have a major problem in that the estimation is inaccurate.
They compute $p_{\theta}(Y \cond \xmi)$ by substituting $\xii$ with other values, but the replaced image is then out-of-distribution (OoD) from the training distribution of a classifier.
Previous studies~\cite{hendrycks2016baseline,liang2017enhancing} discovered that the predicted probability of the classifier rapidly drops for OoD images.
Consequently, the removal of an uninformative feature can lead to a large change in the classifier output, and the feature would falsely receive high attributions.
Inspired by our work, Kim~\etal~\cite{kim2020interpretation} critiqued the same problem in the interpretation methods for natural language processing (NLP) models, and their approach has been accepted in the NLP field.

To overcome the OoD problem, we perform Step 1 by implementing accurate marginalization by employing a powerful generative model.
Consequently, the attribution map becomes more \textit{faithful}~\cite{ribeiro2016should} in terms of a quantitative evaluation metric~\cite{petsiuk2018rise}.
Following the precise estimation in Step 1 and drawing inspiration from information theory, we propose a more theory-grounded quantification scheme in Step 2.
Based on the theoretical background, our method also provides a class independent saliency analysis which is unavailable in the existing methods.
Through a two-fold improvement of a perturbation-based method, we propose two attribution maps---an information gain \textbf{(IG) map} and a point-wise mutual information \textbf{(PMI) map}---which are class-\textit{independent} and class-\textit{specific} explanations, respectively.
Fig.~\ref{fig:example} presents examples.

\section{Background}
\subsection{Attribution method}
Given an input image $\xvec \in \mathbb{R}^{\rm{H} \times \rm{W} \times \rm{C}}$, the classifier predicts its label $y$ ($\rm{H}$, $\rm{W}$, and $\rm{C}$ are the height, width, and the number of channels in the image).
By regarding the unknown label $Y$ as a random variable, the classifier predicts it by calculating a posterior distribution, $p_{\theta}(Y \cond \xvec)$.
For a specific class $y_c$ (typically the classifier's top prediction), to provide a human-interpretable rationale for the prediction $p_{\theta}(y_c \cond \xvec)$, the \textit{attribution method} generates an attribution map, $\mvec \in \mathbb{R}^{\rm{H} \times \rm{W}}$.
An element of $\mvec$, $m_i$, quantifies the contribution of pixel $x_i$ to the prediction; however, its definition and range vary among the different methods.

Starting from the Saliency map~\cite{simonyan2013deep}, backprop-based methods~\cite{bach2015pixel,smilkov2017smoothgrad,springenberg2014striving} including Integrated Gradients~\cite{sundararajan2017axiomatic} and Grad-CAM~\cite{selvaraju2017grad} have been suggested to generate an attribution map using backpropagation.
Although they are sufficiently fast to facilitate real-time results, Dabkowski~\etal~\cite{dabkowski2017real} argued that their quality is limited and suggested the generation of a mask-like attribution map.
Following the work of Dabkowski~\etal~\cite{dabkowski2017real}, mask-based methods including FIDO~\cite{chang2018explaining}, Meaningful Perturbation~\cite{fong2017interpretable}, Extremal Perturbation~\cite{fong2019understanding}, and RISE~\cite{petsiuk2018rise} aim to find a mask that covers a relevant region of the target class.
They applied an image perturbation with a mask and optimized the mask with their unique objective functions.
Meanwhile, Shapley-based methods~\cite{lundberg2017unified,frye2020shapley} take game-theoretic approach to calculate feature importance using Shapley value~\cite{shapley201617}.
More details regarding the backprop-based, mask-based, and Shapley-based methods are provided in the Appendix.

\subsection{Perturbation-based method}
Since the work by Zeiler \etal~\cite{zeiler2014visualizing}, perturbation-based methods define an attribution of an input feature as the difference between the classifier output with and without that feature.
They compute $p_{\theta}(Y \cond \xmi)$ for each feature $\xii$ (Step 1) and measure the discrepancy from the original prediction, $p_{\theta}(Y \cond \xvec)$ (Step 2).
Fig.~\ref{fig:perturbation_methods} depicts the overall flow of the perturbation-based methods.

\begin{figure}[t]
  \begin{center}
    \includegraphics[width=\linewidth]{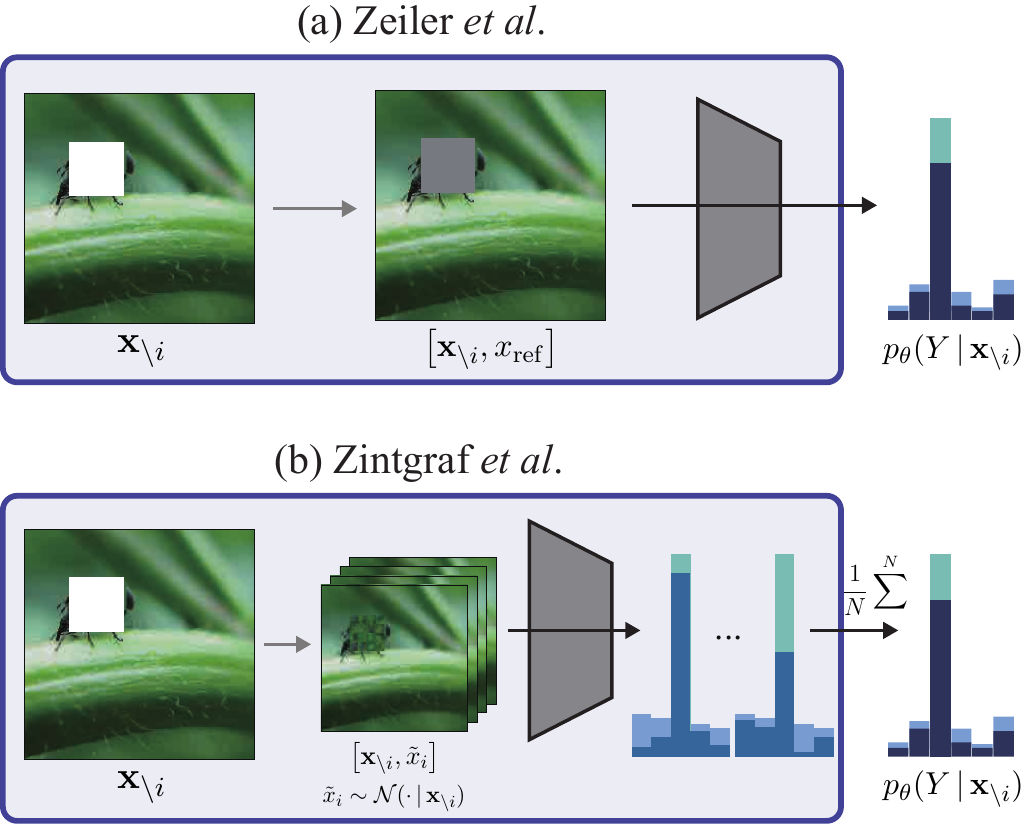}
  \end{center}
  \vspace{-10pt}
  \caption{\textbf{Perturbation-based methods involve estimating the classifier output for a partially removed input, $\xmi$.} Zeiler~\etal~\cite{zeiler2014visualizing} filled the removed pixels with gray value, while Zintgraf~\etal~\cite{zintgraf2017visualizing} used marginalization using Gaussian distribution. }  \label{figure:perturbation_methods_step1} 
\end{figure}

In Step 1, a direct computation of $p_{\theta}(Y \cond \xmi)$ is infeasible because typical classifiers cannot predict for a partially removed input.
Zeiler~\etal~\cite{zeiler2014visualizing} approximated the computation by substituting $x_i$ for a reference value $\xref$:
\begin{equation} \label{eq:reference_value}
    p_{\theta}(Y \cond \xmi) \approx p_{\theta}(Y \cond \xmi, x_{\rm{ref}}),
\end{equation}
where the authors used a gray value for $x_{\rm{ref}}$, as depicted In Fig.~\ref{figure:perturbation_methods_step1}(a).
Although such a heuristic substitution has been widely used \cite{dabkowski2017real,fong2017interpretable,shrikumar2017learning}, selecting the reference value is non-trivial. 
This is because unlike the MNIST dataset, the background of which is black, many datasets (\eg~ImageNet) have no uninformative color; a gray pixel indicates a gray color, not ``no information."
In such cases, a replacement with $x_{\rm{ref}}$ introduces a new flaw, which it leads to an OoD problem.
In Step 2, the authors used $\textrm{Diff}=p_{\theta}(y_c \cond \xvec)-p_{\theta}(y_c \cond \xmi)$, which directly subtracts two probabilities.
The authors' choice was somewhat heuristic because a subtraction of two probabilities has no sound theoretical background.

\begin{figure}[t]
  \begin{center}
    \includegraphics[width=\linewidth]{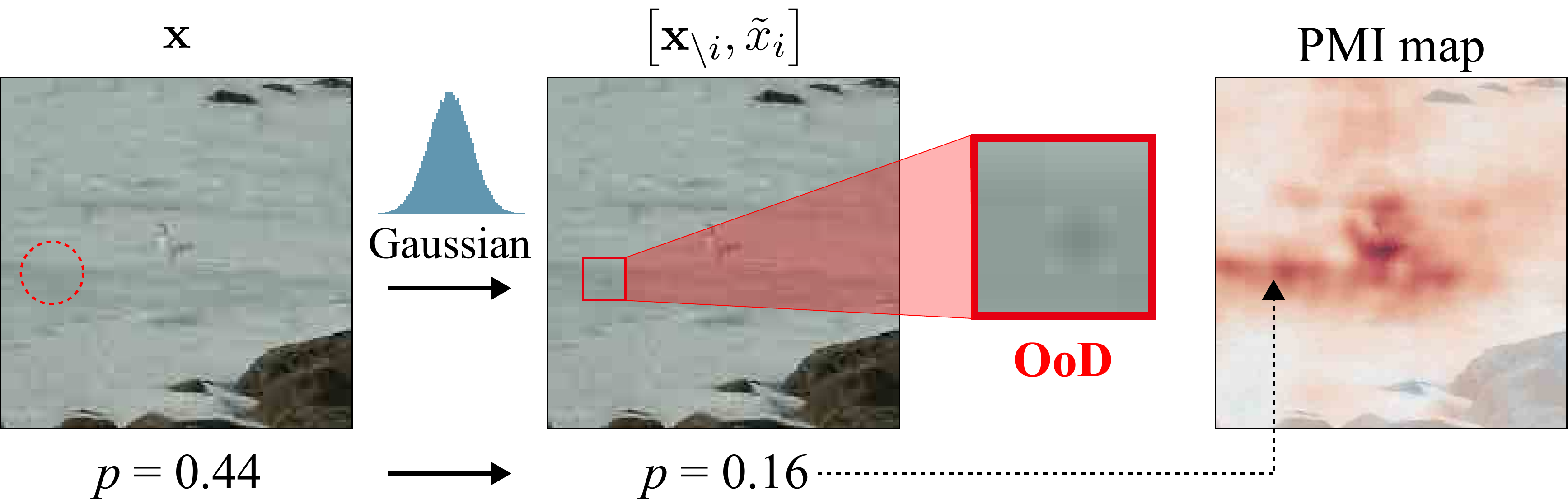}
  \end{center}
  \vspace{-10pt}
  \caption{\textbf{Replacing pixels with values sampled from Gaussian distribution incurs an OoD problem. } The image of a bird is predicted as \textit{ptarmigan} with $p=0.44$ by VGG19~\cite{simonyan2014very}. To simulate the classifier output without a portion of an input (red circle), Zintgraf~\etal~\cite{zintgraf2017visualizing} replace a part of the input using Gaussian distribution.
  When the background pixels are replaced with Gaussian samples, the predicted probability decreases ($p=0.44 \rightarrow 0.16$) even more than when the bird at the center of the image is replaced ($p=0.44 \rightarrow 0.19$). This is due to an OoD problem; the prediction decreases because the image is OoD, not because the background was important. As a consequence, the attribution map cannot distinguish which area is truly important.}  \label{figure:gaussian_ood} 
\end{figure}

An accurate estimation of $p_{\theta}(Y \cond \xmi)$ can be achieved by using marginalization as in (\ref{eq:marginalization}), where multiple values of $\xti$ are sampled from the distribution of $X_i$ (a random variable corresponding to $x_i$).
\begin{equation} \label{eq:marginalization}
    p_{\theta}(Y \cond \xmi) = \mathbb{E}_{\xti \sim p(\Xii \cond \xmi)}\left [p_{\theta}(Y \cond \xmi, \xti) \right ].
\end{equation}
The prediction difference analysis (PDA) \cite{zintgraf2017visualizing} proposed by Zintgraf~\etal~follows this approach with the assumption that $x_i$ follows a Gaussian distribution \ie, $p(\Xii \cond \xmi)=\mathcal{N}(\cdot \cond \xmi)$.
The expectation in (\ref{eq:marginalization}) is further approximated using a Monte Carlo (MC) sampling of sample number $N=10$, as illustrated in Fig.~\ref{figure:perturbation_methods_step1}(b).
However, we show that the Gaussian distribution does not model the image distribution accurately, and thus PDA still suffers from the OoD problem, as in Fig.~\ref{figure:gaussian_ood};
the images substituted by samples from the Gaussian distribution are clearly OoD.
Hence the resulting attribution map cannot distinguish authentic evidence for the classifier.
In Step 2, the authors~\cite{zintgraf2017visualizing} used a weight of evidence: $\textrm{Diff}=\textrm{logodds}(p_\theta(y_c \cond \xvec))-\textrm{logodds}(p_\theta(y_c \cond \xmi))$, where $\textrm{logodds}(p)=\textrm{log}(p/(1-p))$.
By contrast, we utilize point-wise mutual information and information gain, and their definitions and strengths are described in the following section.

\subsection{Information-theoretic analysis} \label{section:information_theory}
Given the random variables $X$ and $Y$ with their realizations $x$ and $y$, information theory discloses how they are interconnected.
Specifically, point-wise mutual information (PMI) between two events is a measure of the extent to which one event \{$X=x$\} provokes the other \{$Y=y$\}: $\textup{PMI}(x;y)=\log(p(y \cond x)/p(y))$.
PMI has a positive or negative value when one event triggers or suppresses another event, respectively.
Therefore, by investigating the sign and magnitude of PMI, we can infer if an observation of \{$X=x$\} is evidence for (positive) or against (negative) event \{$Y=y$\}.

We now consider an event \{$X=x$\} and a random variable $Y$.
If the observation of $x$ brings about a considerable change in the probability distribution of $Y$, the observation is considered informative.
Therefore, we quantify the information as a Kullbeck---Leibler (KL) divergence between the posterior and prior distributions of $Y$ and refer to it as information gain (IG): $\textup{IG}(Y, x) = \textup{D}_\textup{KL}\left(p(Y \cond x) \parallel p(Y) \right)$.
IG is minimized at zero when observing $x$ provides no information regarding $Y$, while a large IG value denotes a significant amount of \textit{Shannon's information}.
Therefore, investigating the magnitude of IG help us estimate the informativeness of an event.

We utilize these two information-theoretic measures, PMI and IG, to quantify the contribution of each feature.

\begin{figure}[t]
  \begin{center}
    \includegraphics[width=\linewidth]{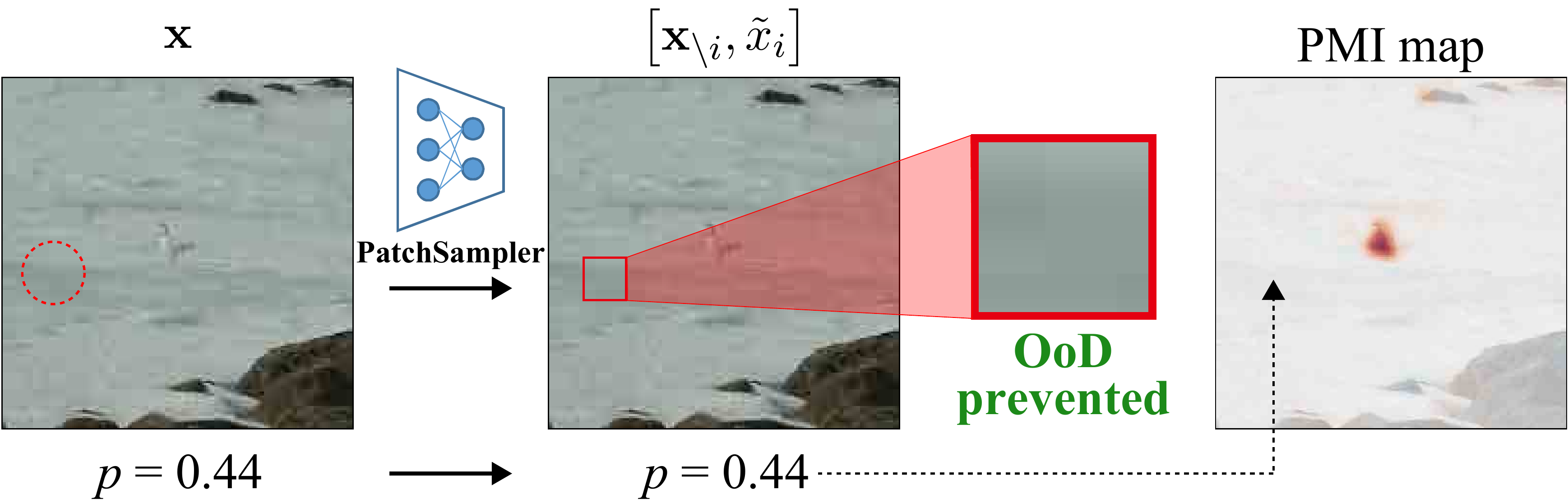}
  \end{center}
  \vspace{-10pt}
  \caption{\textbf{Replacing pixels with PatchSampler does not incur the OoD problem.} As PatchSampler accurately models the distribution of pixels in the image, replacing pixels in the background does not bring a notable change in the classifier output. As a result, the attribution map pinpoints the bird on which the classifier is primarily focused. The improvement in the attribution map is also verified using a quantitative metric, Deletion AUC~\cite{petsiuk2018rise} (the lower the better, and details are in Section~\ref{section:quantitative_comparison}). The attribution map using the proposed PatchSampler yields a lower Deletion AUC (0.002) than using Gaussian (0.007).
  }  \label{figure:enhanced_modeling} 
\end{figure}

\begin{figure*}[t]
    \centering
    \includegraphics[width=0.95\textwidth]{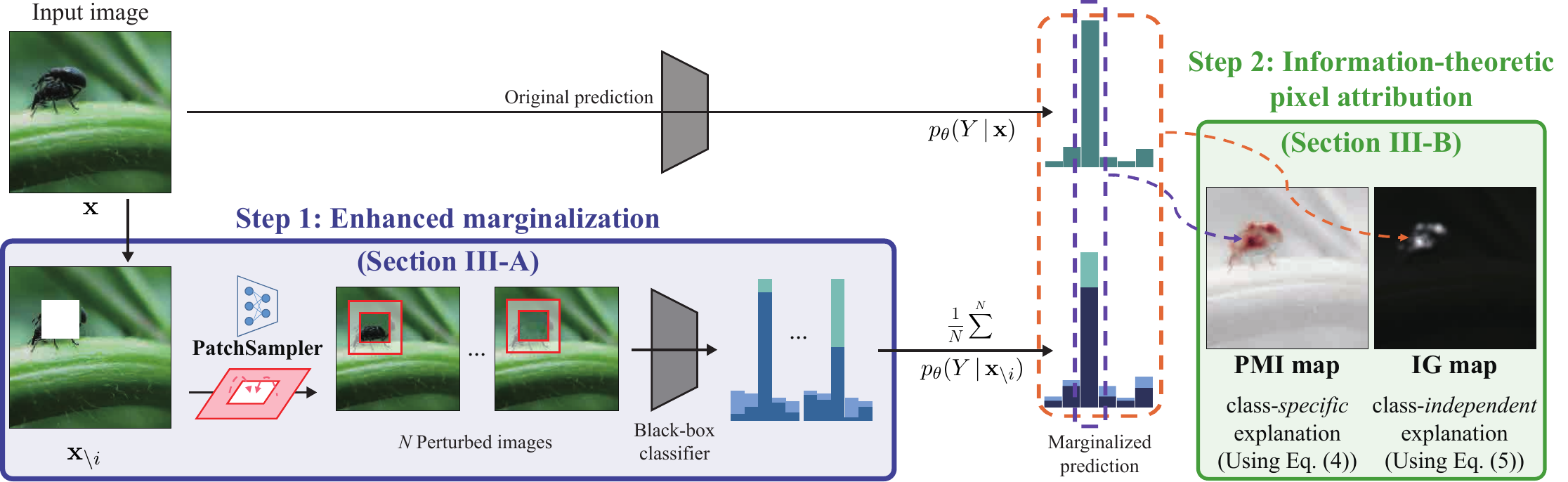}
    \vspace{-10pt}
    \caption{\textbf{The overall flow of the proposed attribution method.}
    For every patch in an image, we compute a marginalized prediction (expected prediction for the image without that patch).
    The difference between the marginalized and original prediction is measured in PMI and IG maps.
    An IG map indicates the \textit{overall} salient area, and a PMI map suggests the region for and against a \textit{specific class}.
    }
    \label{fig:big_picture}
\end{figure*}

\section{Method}
In this study, we propose a model-agnostic \textit{perturbation-based} method.
The method visually explains the prediction of any black-box classifier using two building blocks---enhanced marginalization (Section~\ref{section:marginalization}) and information-theoretic pixel attribution (Section~\ref{section:pixel_attribution}).
Each component corresponds to Step 1 and Step 2 in the perturbation-based methods.

\subsection{Enhanced marginalization} \label{section:marginalization}
To perform the marginalization expressed in (\ref{eq:marginalization}), we should sample input features, and thus a generative model $p(\Xii \cond \xmi)$ is necessary.
If the generative model does not model the distribution correctly, it makes the perturbed images ($\left[\xmi, \xti\right]$ in (\ref{eq:marginalization})) OoD.
Chang \etal~\cite{chang2018explaining} partially addressed the OoD problem by replacing inputs using the CA-GAN \cite{yu2018generative} output, namely, they used $\xref=\textrm{CA-GAN}(\xmi)$ for computing (\ref{eq:reference_value}).
Nonetheless, their approach remains far from marginalization because CA-GAN in-fills the input deterministically without sampling.

To model the complex distribution of pixels and sample multiple pixel values, we deploy a powerful generative model consisting of a neural network, referred to as ``PatchSampler".
Because a pixel is strongly dependent on its surroundings, PatchSampler approximates $p(\Xii \cond \xmi) \approx p(\Xii \cond \xhmi)$, where $\xhmi$ is the neighborhood of the pixel $\xii$.
$p(\Xii \cond \xhmi)$ is then modeled $p_\phi(\Xii \cond \xhmi)$ using a neural network comprising a series of convolutional layers with parameter $\phi$.
We trained PatchSampler in advance using the training data of the classifier.
Compared to Gaussian in Fig.~\ref{figure:gaussian_ood}, PatchSampler models the distribution far more accurately, as shown in Fig.~\ref{figure:enhanced_modeling}.
As a result, the computation in Step 1 becomes more accurate, and the OoD problem is prevented.
The resulting attribution maps depicted in Fig.~\ref{figure:enhanced_modeling} show a dramatic improvement both qualitatively and quantitatively.
As the PatchSampler was trained
using the ImageNet training set, which is a large general image dataset, the trained PatchSampler is expected to apply to classifiers for the other natural image datasets.

Evaluating (\ref{eq:marginalization}) requires a classifier feedforward $p_{\theta}(Y \cond \xmi, \xii)$ for all possible values of $\xii$, which demands a large number of computations.
We mitigate this computing issue by using the MC approximation of the sample number $N$:
\begin{equation} \label{eq:ours}
    p_{\theta}(Y \cond \xmi) \approx \frac{1}{N}\sum_{\xti \sim p_\phi(\cdot \cond \xhmi)}^{N}p_{\theta}(Y \cond \xmi, \xti).
\end{equation}
Although a larger $N$ yields better approximation, we empirically show that a small $N$ provides visually and numerically indistinguishable results in the Appendix.

Given a class $y_c$, an input feature $\xii$ obtains a high attribution when the corresponding $p_{\theta}(y_c \cond \xmi)$ is far from $p_{\theta}(y_c \cond \xvec)$.
This happens under following two circumstances.
First, $\xii$ should be relevant to class $y_c$.
Second, the probable values of $\Xii$ should be diverse, so that observing $\{ \Xii=\xii \}$ makes the posterior different from the prior.
In other words, pixels receive high attribution when they are both \textit{relevant} to $y_c$ and \textit{unpredictable} \cite{zintgraf2017visualizing}.
However, pixels are highly correlated with their neighborhoods and are extremely predictable.
Alternatively, patches are more uncertain and less predictable.
Since a pixel is a patch of unit size, we regard the contribution of a patch instead of a pixel for a generality.
We use $\xii$ a patch of size $K \times K$ with a hyperparameter $K$.
Accordingly, PatchSampler models a joint distribution of a $K \times K$ patch conditioned on its surrounding $3K \times 3K$ patch.
For $K > 1$, patch-wise calculated attributions were equally distributed to every pixel in each patch.

Throughout the research, we used $K=8$ as it provides easily understandable attribution maps.
However, the most attractive value of $K$ is different for each input image.
In the Appendix, we provide an example of the attribution maps generated using various values of $K$.
One can try various values of $K$ to obtain more intuitive visualization depending on the size of the object.

\subsection{Information-theoretic pixel attribution} \label{section:pixel_attribution}
Given an image $\xvec$ and its (unknown) label, we consider their corresponding random variables, $\Xvec$ and $Y$.
The classifier prediction $p_{\theta}(Y \cond \xvec)$ is the posterior distribution of $Y$.
Starting from a prior distribution, the classifier makes a final prediction based on the information from the input observation.
To quantify the contribution of the features, it is reasonable to measure the amount of information provided by each feature.
Accordingly, our method outputs two attribution maps, $\pmimap$ (PMI map) and $\igmap$ (IG map) $\in \mathbb{R}^{\rm{H} \times \rm{W}}$, for two different questions: ``How much does each pixel support a specific class $y_c$?" and ``How informative is each pixel?"


\subsubsection{PMI map}
First, assume that we are interested in how much pixel $\xii$ accounts for class $y_c$.
In other words, given the other part of the image, how much does an observation \{$\Xii=\xii$\} trigger the predicted event \{$Y=y_c$\}?
Such a notion is captured by PMI between the two events conditioned on $\xmi$.
For a given pixel, its PMI for a class $y_c$ is computed as follows:
\begin{equation} \label{eq:pmi}
    \textup{PMI}(y_c;\xii \cond \xmi)=\log \left (\frac{p_\theta(y_c \cond \xmi, \xii)}{p_\theta(y_c \cond \xmi)} \right).
\end{equation}
It is noteworthy that the numerator in (\ref{eq:pmi}) is the prediction for the original input.
For every pixel, we calculate each PMI value using (\ref{eq:pmi}) and the result constitutes a \textbf{PMI map}: $\pmimap^{i}(y_c,\xvec) \defeq \textup{PMI}(y_c;\xii \cond \xmi)$.
Pixels with positive PMI support $y_c$, whereas those with negative PMI oppose $y_c$.
Note that a PMI map can be calculated for any class $y_c$, not necessarily the top-1 class.

\subsubsection{IG map}
To quantify the informativeness of a pixel $\xii$ regardless of a specific class,
we estimate the IG between an observation \{$\Xii=\xii$\} and $Y$ given $\xmi$ using (\ref{eq:ig}).
\begin{align*} \label{eq:ig}
    \textup{IG}(Y, \xii \cond \xmi) &= \kl \left (p_\theta(Y \cond \xmi, \xii) \parallel p_\theta(Y \cond \xmi) \right) \\
    &= \mathbb{E}_{y_c \sim p_\theta(Y \cond \xvec)}\left [\log \left (\frac{p_\theta(y_c \cond \xmi, \xii)}{p_\theta(y_c \cond \xmi)} \right) \right ] \\
    &= \numberthis \mathbb{E}_{y_c \sim p_\theta(Y \cond \xvec)}\left [\textup{PMI}(y_c;\xii \cond \xmi)\right ].
\end{align*}
Likewise, the \textbf{IG map} is defined as $\igmap^{i}(\xvec) \defeq \textup{IG}(Y, \xii \cond \xmi)$.
Notably, the IG is an expectation of the PMI among all possible $y_c$ in (\ref{eq:ig}).
This allows the calculation of IG by obtaining the PMI for every class $y_c$ and calculating the weighted sum with the predicted probability $p_\theta(y_c \cond \xvec)$.
Because a single feed-forward of a classifier calculates $p_\theta(y_c \cond \xvec)$ for every $y_c$, an IG map can be obtained at only a marginal cost given PMI maps.
Pixels with high IG attribution are informative to the classifier and thus salient.

The IG map is not class-specific and is complementary to a class-specific PMI map.
In particular, when multiple classes are probable so that no class-specific explanations are sufficient to describe the behavior of the classifier, only the IG map can provide an acceptable interpretation.
Note that such class-independent explanations are not available in most existing methods, including perturbation-based methods~\cite{zeiler2014visualizing,zintgraf2017visualizing}.
In the calculation of Eq~(\ref{eq:pmi}) and (\ref{eq:ig}), we add a small constant ($\epsilon$=1e-13) inside log for the numerical stability.

In summary, our method generates two attribution maps, namely a PMI map and an IG map, which provide class-specific and class-independent explanations, respectively.
Fig.~\ref{fig:big_picture} shows an overview of the proposed method.
The pseudo-code of our method is provided in the Appendix.

\begin{figure}[t]
  \begin{center}
    \includegraphics[width=\linewidth]{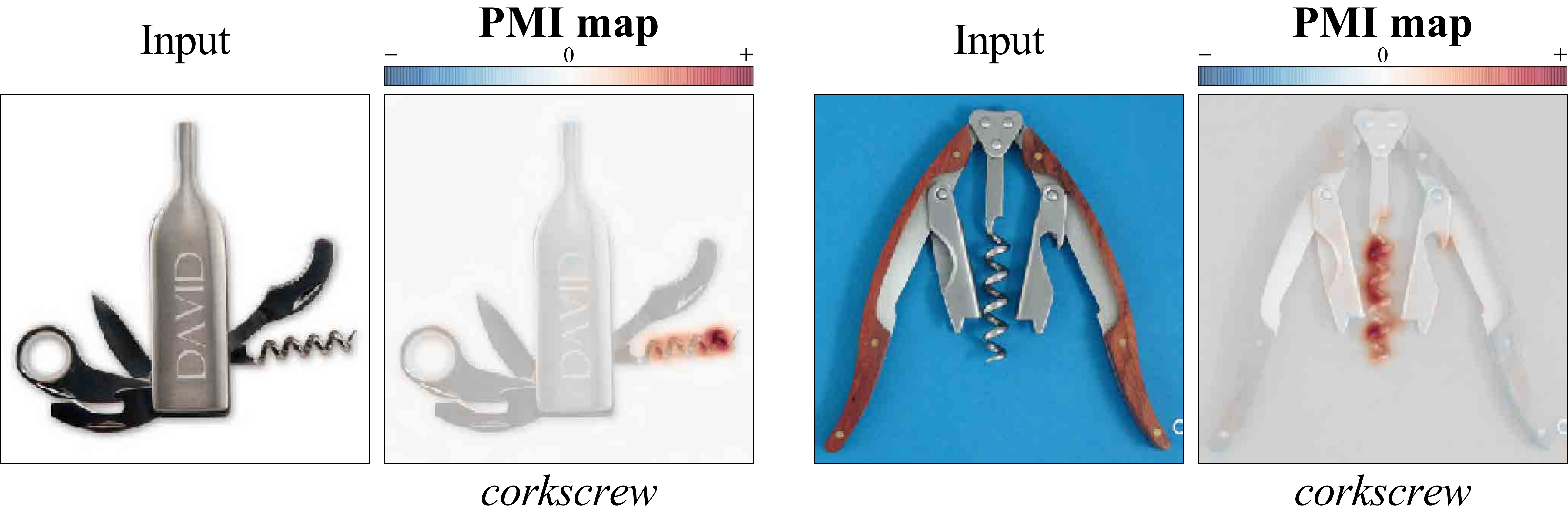}
  \end{center}
  \vspace{-10pt}
  \caption{\textbf{PMI maps for two corkscrew images.} The PMI map for each image shows that the corkscrew in the image is the reason for the classification results.}  \label{figure:PMImap_ex} 
\end{figure}

\begin{figure}[t]
  \begin{center}
    \includegraphics[width=\linewidth]{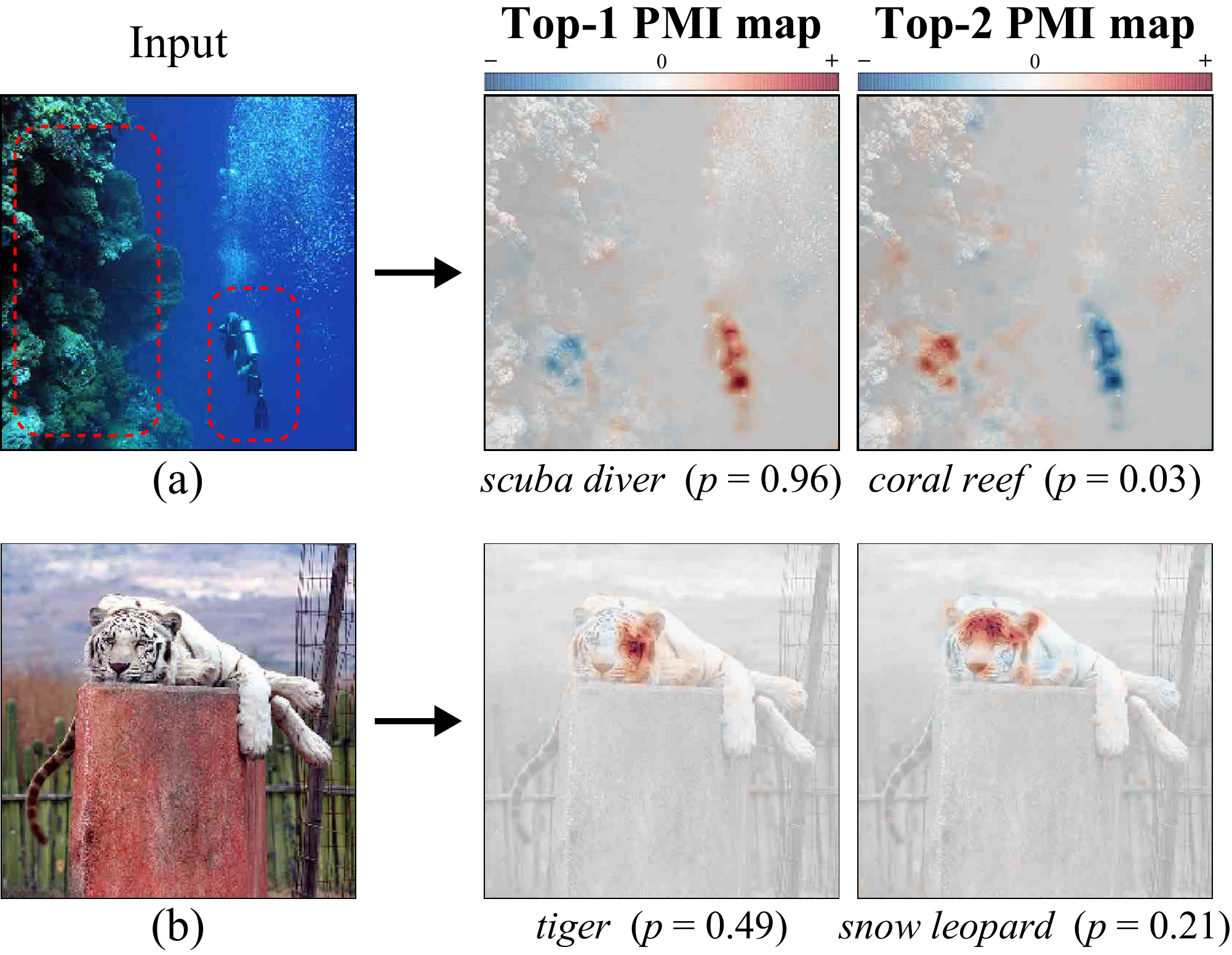}
  \end{center}
  \vspace{-10pt}
  \caption{\textbf{PMI maps for the top-1 and top-2 predicted classes.} In (a), the PMI map shows the opposite influences of two objects (scuba diver and coral reef) to the corresponding classes. In (b), the facial area of the snow leopard supports both top-1 and top-2 classes.}
  \vspace{-10pt}
  \label{fig:pmi2_scubadiver}
\end{figure}

\subsubsection{Comparison to Learning to Explain~\cite{l2x}}
Inspired from an information-theoretic perspective, Learning to Explain (L2X) tries to explain a prediction by selecting a set of informative input features.
L2X trains a selector model that predicts a set of features with a high variational lower bound of mutual information.
Unlike PDA~\cite{zintgraf2017visualizing} and our method, L2X does not consider the distribution of a feature and uses a variational approximation with a heuristic black replacement.
For this reason, MNIST was the only image dataset that the authors demonstrated.
Taking the pixel distribution into account, we attempt to explain a classifier trained with datasets of any kind.
Moreover, we directly quantify the information of each feature so that our method can deduce a solution for their problem, but not vice versa.

\section{Results and Discussion}
First, we present the interpretation results of a classifier prediction using the proposed method.
Second, we compare the proposed method both quantitatively and qualitatively to the existing methods.
Third, we present the strengths of the proposed method.
Finally, we analyze the proposed method in detail.
More results regarding the hyperparameters and the implementation details are provided in the Appendix.
Throughout the experiments, we interpreted the predictions of VGG19~\cite{simonyan2014very} classifier for various images.
Besides this specific classifier, our method applies to any other model, and the explanation results for other image classifiers are also provided in the Appendix.

\begin{figure}[t]
    \centering
    \includegraphics[width=\linewidth]{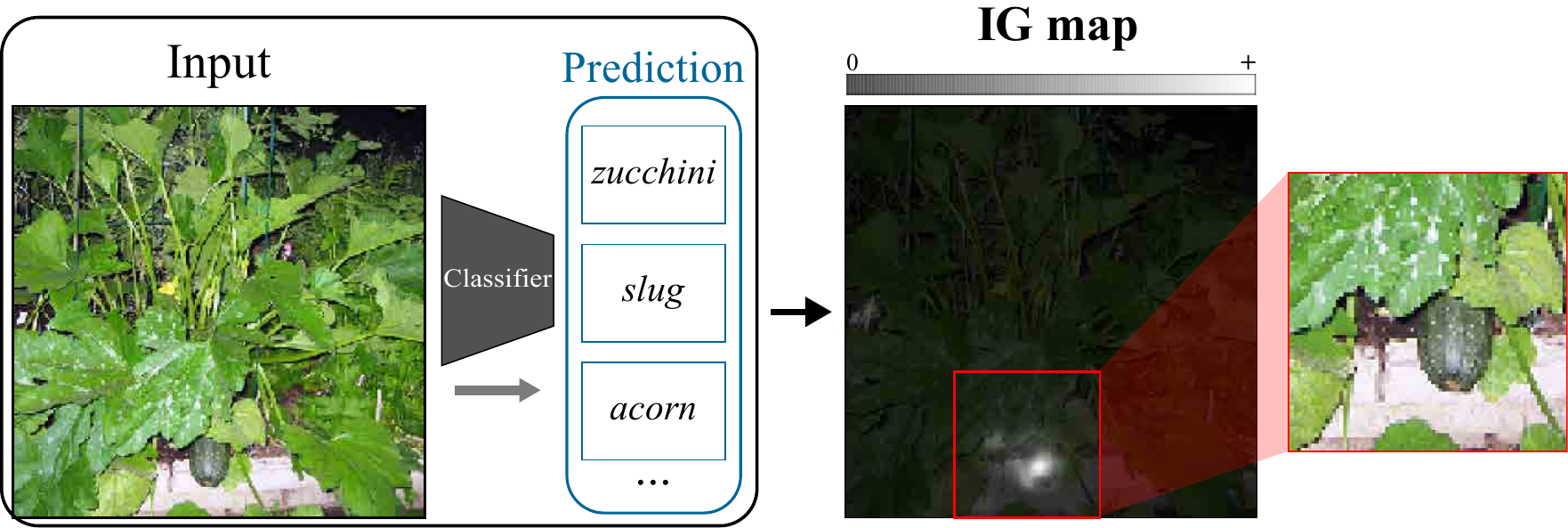}
    \vspace{-15pt}
    \caption{
    \textbf{IG map highlights where the classifier mostly focused on.} VGG19~\cite{simonyan2014very} failed to make a confident prediction. The IG map shows that the classifier was trying to classify the herb at the bottom.
    }
    \label{fig:useful_igmap}
    \vspace{-5pt}
\end{figure}

\begin{figure*}[t]
    \centering
    \includegraphics[width=\textwidth]{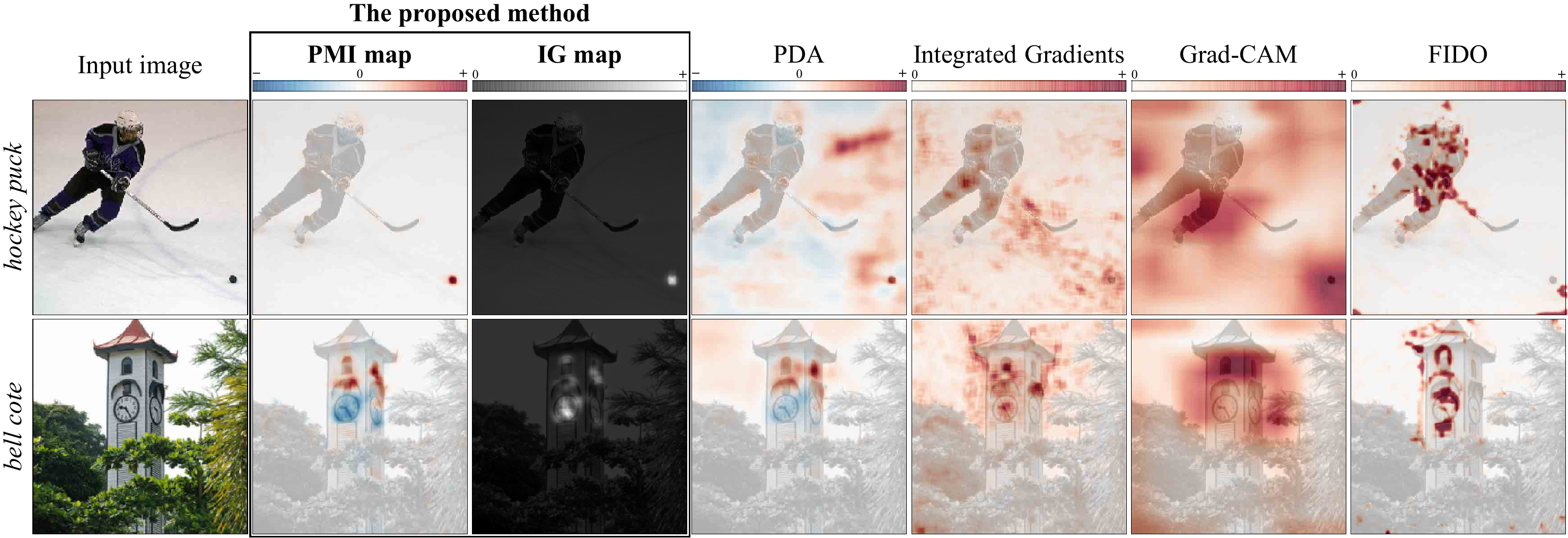}
    \vspace{-10pt}
    \caption{
    \textbf{Explanations for VGG19~\cite{simonyan2014very} predictions.}
    We compare PMI and IG maps with the other attribution maps generated by popular methods in each category (PDA~\cite{zintgraf2017visualizing}, Integrated Gradients~\cite{sundararajan2017axiomatic}, Grad-CAM~\cite{selvaraju2017grad}, and FIDO~\cite{chang2018explaining}).
    Integrated Gradients maps were smoothed using a flat kernel of size $K=8$ (the same $K$ used in the PDA, PMI and IG maps) due to their low visibility.
    }
    \vspace{-10pt}
    \label{fig:comparison_all}
\end{figure*}

\subsection{Interpretation results} \label{section:interpretation_results}
\subsubsection{PMI maps}
In Fig.~\ref{figure:PMImap_ex}, we provide the examples of the PMI maps for the classifier prediction.
The images are classified as \textit{corkscrew}, and their PMI maps clearly indicate that the corkscrews are the reason why.
When an image contains multiple objects, PMI map of each class can visualize the influence of each object on the corresponding class.
For example, the PMI maps in Fig.~\ref{fig:pmi2_scubadiver}(a) suggest that the scuba diver supports the \textit{scuba diver} class, but contradicts the rival class, \textit{coral reef}, and vice versa.
Meanwhile, evidence for each class is not always mutually exclusive, particularly when the classes are similar;.
In Fig.~\ref{fig:pmi2_scubadiver}(b), the face of the snow leopard is supporting evidence for both \textit{tiger} and \textit{snow leopard} classes.

\subsubsection{IG maps}
In Fig.~\ref{fig:useful_igmap}, the classifier failed to pick a dominant class; the top-3 predicted classes were \textit{zucchini} ($p=0.20$), \textit{slug} ($p=0.18$), and \textit{acorn} ($p=0.14$).
In such cases, class-specific attribution maps are hardly helpful, and class-independent explanations produced by the IG map are the only remaining option.
In Fig.~\ref{fig:useful_igmap}, the IG map indicates that the classifier was trying to classify the herb at the bottom.

\begin{figure}[t]
  \begin{center}
    \includegraphics[width=\linewidth]{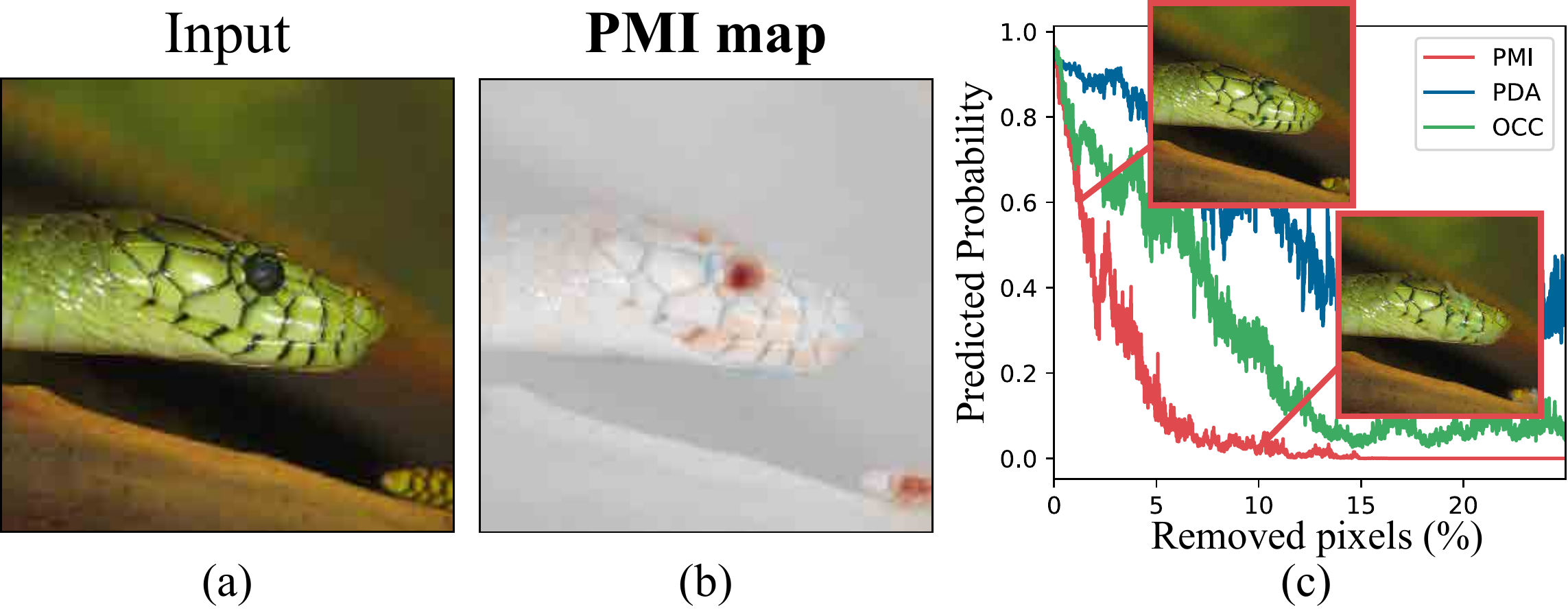}
  \end{center}
  \vspace{-5pt}
  \caption{\textbf{An input image (a), the corresponding PMI map (b), and the deletion curves for the image (c).} A sharp drop in the deletion curve (c) signifies a faithful attribution map. In (c), two example images masked out using CA-GAN~\cite{yu2018generative} show that removing the eye, which is highlighted by the PMI map, quickly hurts the prediction.
  }
  \label{fig:deletion_curves}
\end{figure}

\begin{figure}
    \centering
    
    \includegraphics[width=\linewidth]{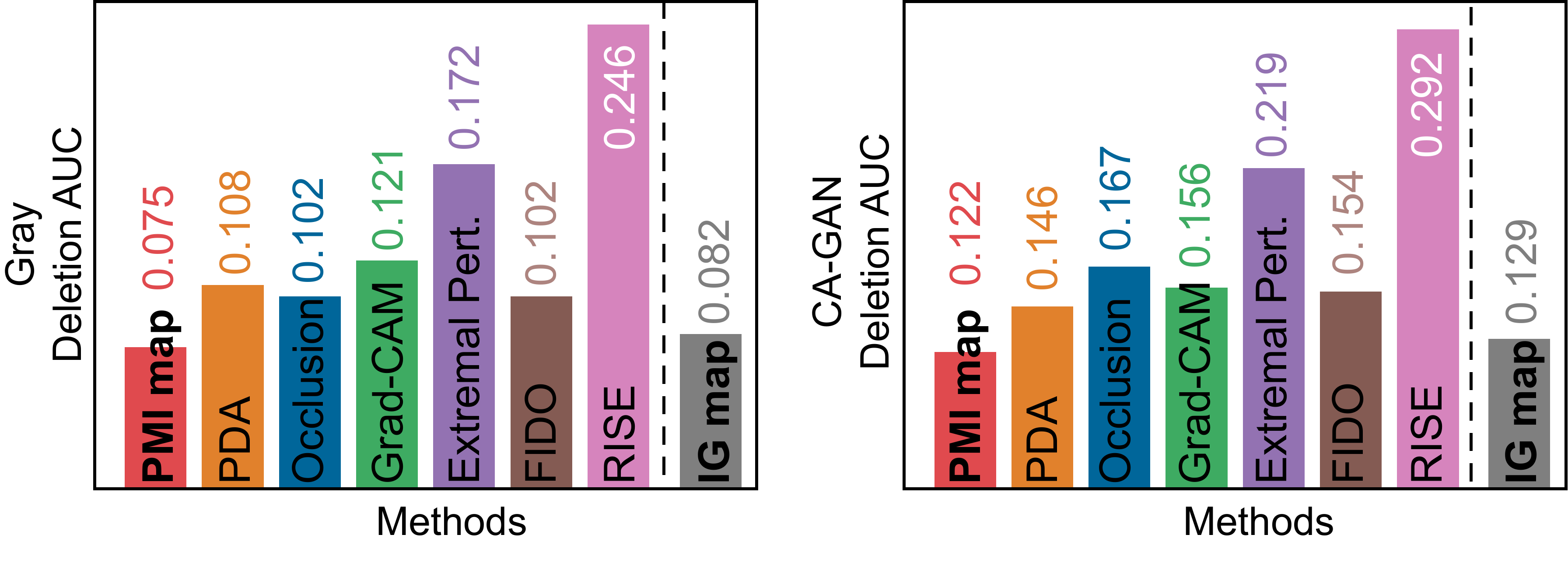}
    \vspace{-20pt}
    \caption{\textbf{Average Deletion AUCs~\cite{petsiuk2018rise} for 5,000 random ImageNet test images.} The PMI map yields the lowest Deletion AUC both in gray removal (left) and CA-GAN~\cite{yu2018generative} in-fill (right) experiments.}
    \label{fig:deletion_scores}
    
\end{figure}

\begin{figure}
    \centering
    \includegraphics[width=0.75\linewidth]{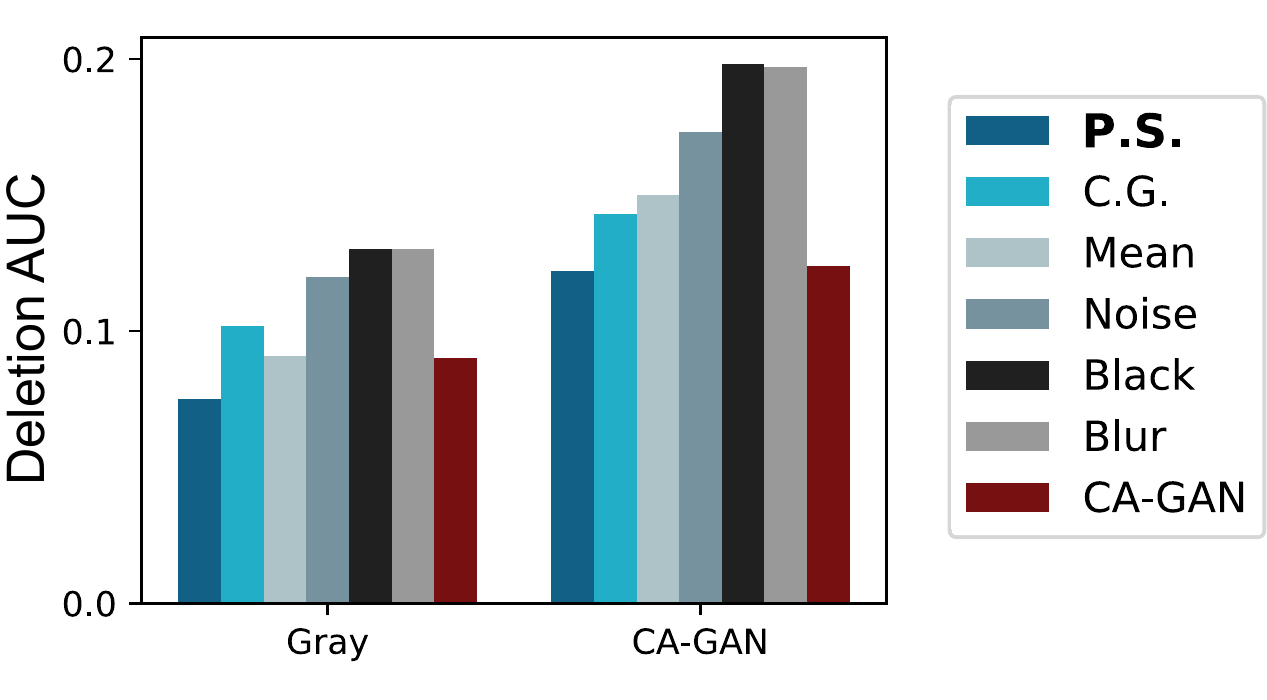}
    \vspace{-5pt}
    \caption{\textbf{Average Deletion AUCs~\cite{petsiuk2018rise} of PMI maps for 5,000 random ImageNet test images generated using various generative models and heuristic in-fill methods.} Using PatchSampler (P.S.) as a generative model yields lower Deletion AUC~\cite{petsiuk2018rise} than using conditional Gaussian (C.G.) and the other heuristic replacements (replacement with a patch mean, noise, or black).}
    \label{fig:deletion_scores_generative}
\end{figure}

\subsection{Comparison to the existing methods} \label{section:comparison}

\subsubsection{Quantitative comparison} \label{section:quantitative_comparison}
In this section, we examine the \textit{faithfulness}~\cite{ribeiro2016should} of our attribution method through a quantitative analysis.
A faithful explanation is one that correctly explains the behavior of the classifier~\cite{ribeiro2016should}.
Deletion AUC~\cite{petsiuk2018rise}, the metric to be used, measures the area under the prediction probability curve as pixels with high attributions are gradually removed.
A low AUC implies a steep decrease in prediction, thus indicating that the explanation correctly captures the relevant area for the predicted class.
To gradually remove pixels, the authors in \cite{petsiuk2018rise} masked them using gray values. In addition to this, we tried in-filling the pixels with CA-GAN~\cite{yu2018generative}. For example, given an image in Fig.~\ref{fig:deletion_curves}(a), the PMI map for the \textit{green mamba} class is provided in Fig.~\ref{fig:deletion_curves}(b).
Fig.~\ref{fig:deletion_curves}(c) shows the predicted probability curve as the pixels with high scores are gradually removed.
In this example, the deletion curve of the PMI map shows the steepest drop, and hence the PMI map provides the explanation with the highest faithfulness.



Deletion AUCs~\cite{petsiuk2018rise} are compared between maps from closely related methods (perturbation-based and mask-based methods) and are presented in Fig.~\ref{fig:deletion_scores}.
The PMI maps yield the lowest average Deletion AUCs both in gray and CA-GAN~\cite{yu2018generative} in-fill experiments.
The results confirmed the improved faithfulness of the PMI map over the perturbation-based baselines.
As the IG map is a class-independent attribution map, it does not necessarily yield a low AUC.
Interestingly, the AUC for the IG map is quite low, and we postulate that this is because the regions that discriminate among the most probable classes deliver significant information to the classifier overall as well.

\subsubsection{PatchSampler} \label{section:patchsampler_result}

\begin{figure}[t]
  \begin{center}
    \includegraphics[width=\linewidth]{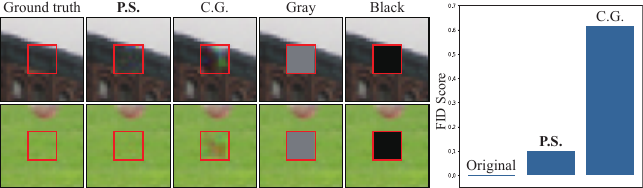}
  \end{center}
  \vspace{-10pt}
  \caption{\textbf{Samples from PatchSampler (P.S.) are more realistic.}
  Samples (left) from the other methods clearly indicate the OoD problem.
  Perturbed images by PatchSampler yield a much lower FID score~\cite{heusel2017gans} (right) than conditional Gaussian (C.G.) from~\cite{zintgraf2017visualizing}.
  }\label{figure:patches}
\end{figure}

The key contribution of this study is the deployment of PatchSampler to accurately perform marginalization in Step 1.
The Deletion AUCs presented in Fig.~\ref{fig:deletion_scores_generative} show that adopting PatchSampler as a generative model lowers both Deletion AUC metrics compared to using heuristic values.
We can conclude that the use of PatchSampler improves the correctness of the attribution maps.
When other type of generative model, CA-GAN~\cite{yu2018generative}, is used, the Deletion AUCs slightly increase.
CA-GAN models the image distribution quite well and prevents the OoD problem, but it cannot accurately implement Eq~(\ref{eq:marginalization}) because it cannot sample multiple pixel values for marginalization.
Moreover, Fig.~\ref{figure:patches} presents the example patches substituted by baseline methods and PatchSampler, and it is evident that the patches replaced by PatchSampler appear more natural.
When quantitatively compared, PatchSampler yields a much lower FID score~\cite{heusel2017gans} (the lower the better) than the Gaussian used in PDA~\cite{zintgraf2017visualizing}.
From the above experimental evidences, our purpose to keep the perturbed images in-distribution seems to be achieved by adopting PatchSampler.

One might suspect that explaining a black-box model by using PatchSampler, which is another black-box model, hurts the trustworthiness of the explanation.
However, we can easily inspect the samples from PatchSampler and understand their impact on the attribution calculation. Therefore, we believe that the explanation provided by the PMI and IG maps confined the un-interpretability to the acceptable level.

\begin{figure}
    \centering
    
    \includegraphics[width=\linewidth]{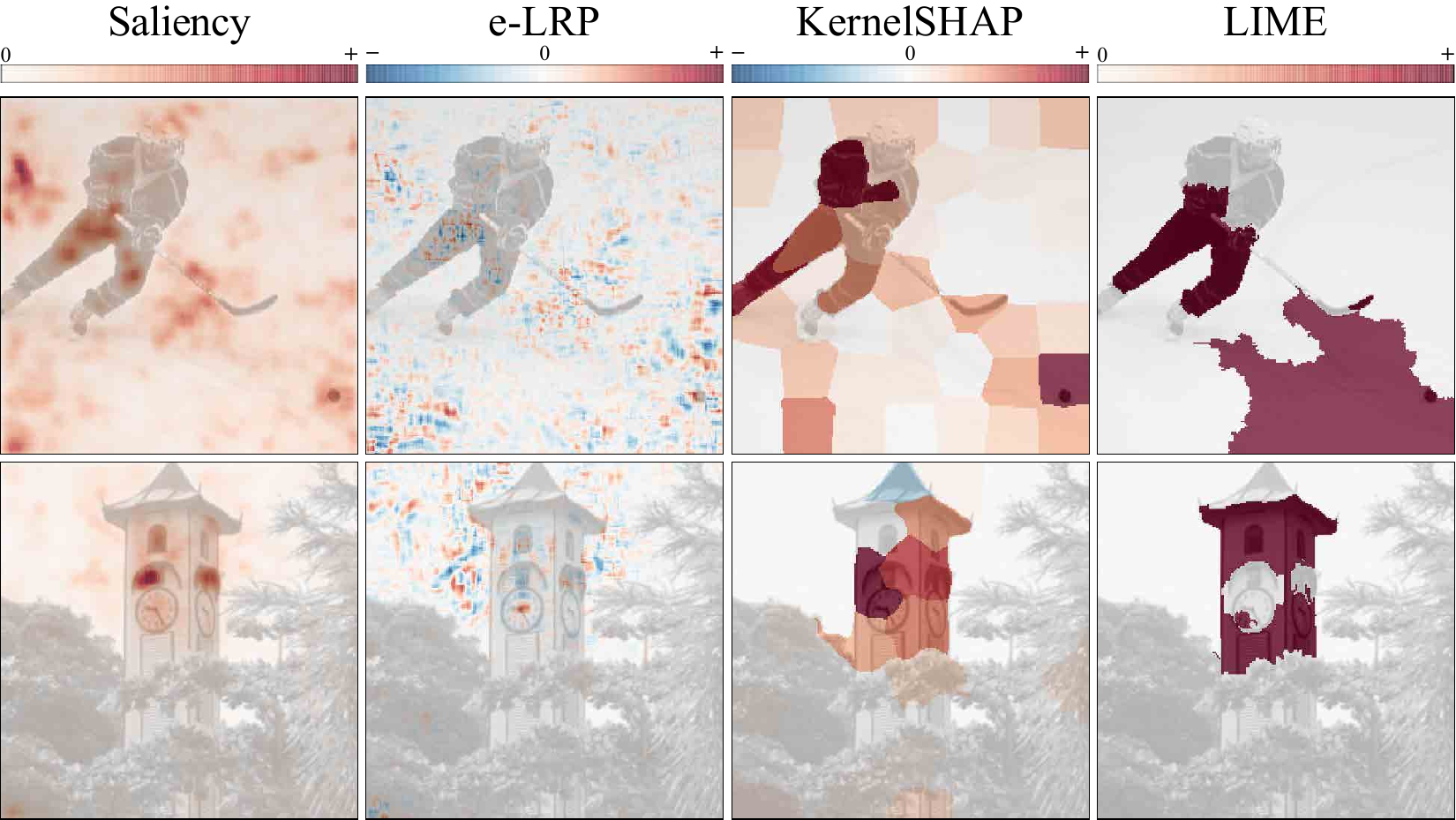}
    \vspace{-20pt}
    \caption{\textbf{Explanations for VGG19~\cite{simonyan2014very} predictions.}
    Various existing attribution methods (the Saliency map~\cite{simonyan2013deep}, e-LRP~\cite{bach2015pixel}, KernelSHAP~\cite{lundberg2017unified}, and LIME~\cite{ribeiro2016should}) explain the prediction of the classifier for the same images in Fig.~\ref{fig:comparison_all}.
    Backprop-based method such as the Saliency map and e-LRP are smoothed using a flat kernel of size $K=8$ due to their low visibility.}
    \label{fig:other_comparison}
    
\end{figure}

\subsubsection{Visual assessment} \label{section:visual_comparison}
We present two example images and attribution maps for the corresponding classifier predictions in Fig.~\ref{fig:comparison_all}.
In the hockey puck image (Fig.~\ref{fig:comparison_all} top), the proposed PMI and IG maps clearly identify the relevant object, the hockey puck.
Compared with our method, PDA~\cite{zintgraf2017visualizing} assigns excessive attribution to irrelevant areas such as the background.
This is due to the OoD problem incurred by the Gaussian modeling of images.

A bell cote image (Fig.~\ref{fig:comparison_all} bottom) is an example showing that the PMI map can offer negative evidence.
The PMI map pinpoints the bell as a supporting evidence, while the clock as negative.
It transpired that the clock supports another class, \textit{analog clock}.
Note that the other methods except PDA~\cite{zintgraf2017visualizing}, which is a perturbation-based method, only visualize supporting evidence.

\begin{figure}[]
  \begin{center}
    \includegraphics[width=\linewidth]{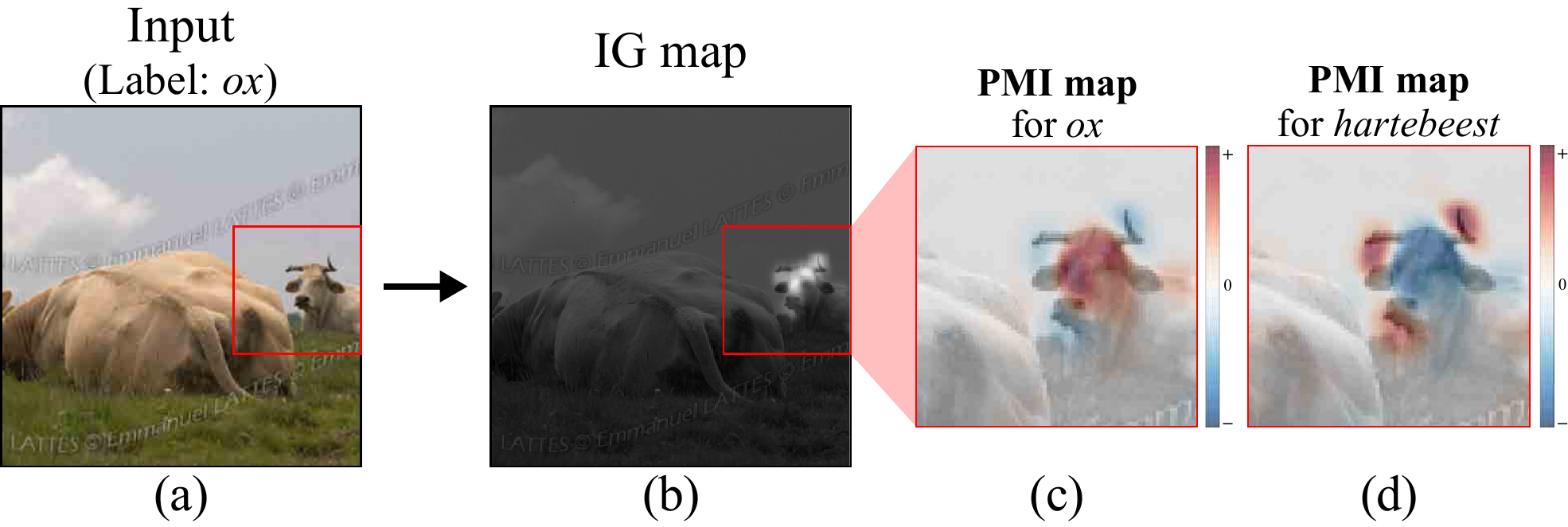}
  \end{center}
  \vspace{-10pt}
  \caption{\textbf{Negative evidence explains why the classifier rejected a class.} VGG19~\cite{simonyan2014very} mis-classified the \textit{ox} image (a) as a \textit{hartebeest}.
  The PMI map for \textit{ox} (c) implies that by observing its ox-like face, the image could be classified as an \textit{ox}, but the horns rejected it.
  The PMI map for \textit{hartebeest} (d) shows that the horns support the \textit{hartebeest} class rather than the \textit{ox}.
  This is because all hartebeests have horns in ImageNet training data, but not all oxen.}
  \label{fig:negative_evidence}
  \vspace{-5pt}
\end{figure}

\subsubsection{Comparison to the methods in other categories}
In this section, we present a comparison with the attribution methods other than the methods compared in Section~\ref{section:comparison}.
In Fig.~\ref{fig:other_comparison}, we provide the attribution maps generated using the Saliency map~\cite{simonyan2013deep}, e-LRP~\cite{bach2015pixel}, KernelSHAP~\cite{lundberg2017unified}, and LIME~\cite{ribeiro2016should} for the same images in Fig.~\ref{fig:comparison_all}.
Backprop-based methods generate extremely scattered attribution map and provide low visibility.
KernelSHAP~\cite{lundberg2017unified} and LIME~\cite{ribeiro2016should} use super-pixels as the unit of perturbation, and hence their attribution maps have a block-like structure.
\setlength{\columnsep}{8pt}%
\begin{wrapfigure}{r}{0.5\linewidth}
  \vspace{-15pt}
  \begin{center}
    \includegraphics[width=\linewidth]{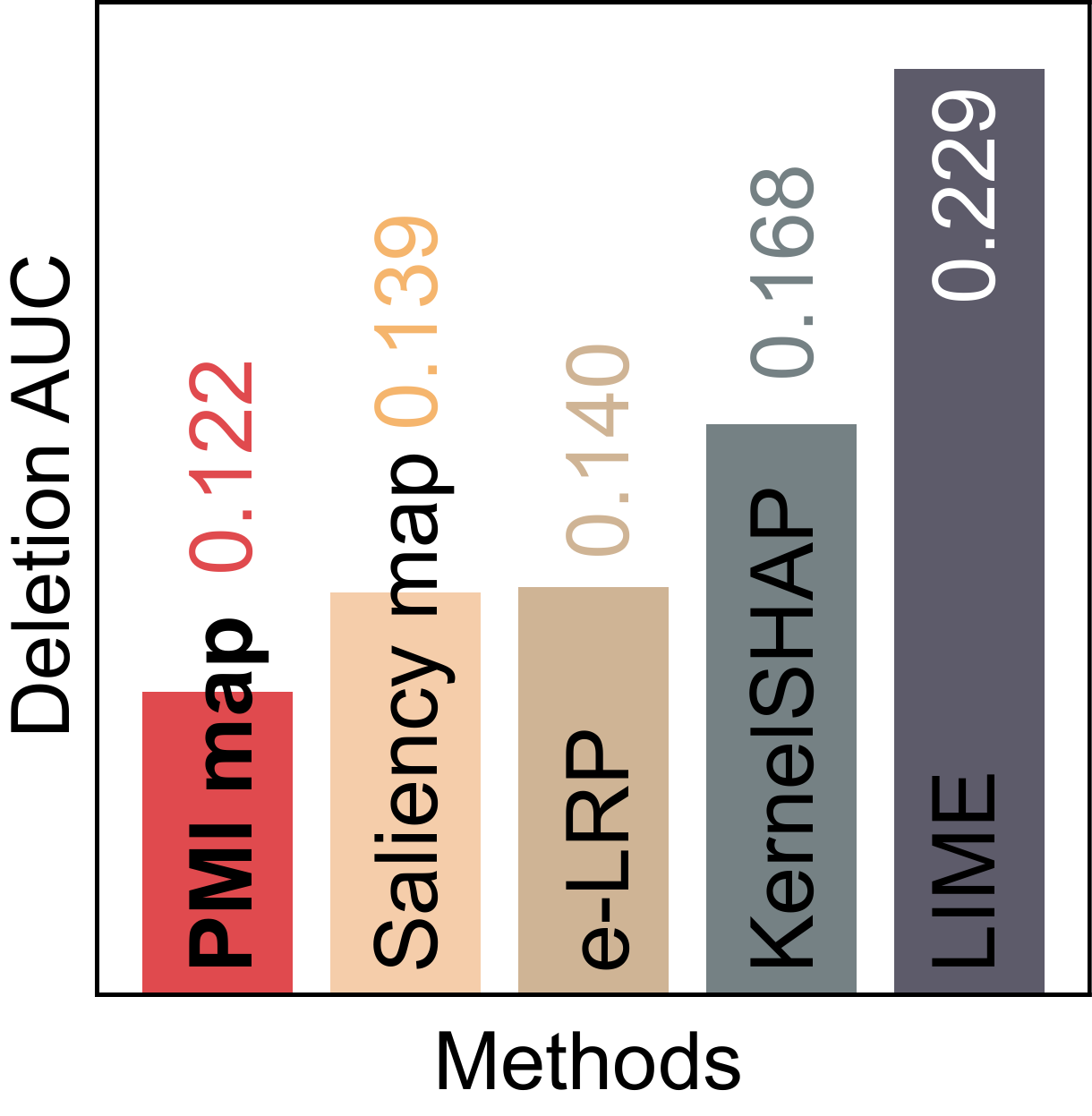}
  \end{center}
  \vspace{-18pt}
  \caption{\textbf{CA-GAN Deletion AUCs.}} \label{fig:deletion_other}
  \vspace{-20pt}
\end{wrapfigure}
They highlighted the key objects in the images (the hockey puck and the bell) with redundant regions.
Quantitatively compared, the average CA-GAN~\cite{yu2018generative} Deletion AUCs~\cite{petsiuk2018rise} for the random 5,000 ImageNet images in Fig.~\ref{fig:deletion_other} show that the baseline methods result in higher AUC than the PMI map, meaning that their maps indicate the evidence of the classifier less accurately than our method.

\subsection{Strengths}

\subsubsection{Negative evidence} \label{section:negative_evidence}

Explanations become more descriptive when they provide contrary as well as supporting evidence.
In Fig.~\ref{fig:negative_evidence}, the PMI map explains the rationales behind the prediction using negative evidence.
Negative evidence facilitates a more in-depth understanding of a classifier.
However, mask-based and most backprop-based methods do not support such analysis because they only provide supporting regions.
To verify whether the negative evidence truly has a negative influence on the classifier prediction, we conducted the following experiment.

\begin{figure}[t]
  \begin{center}
    \includegraphics[width=\linewidth]{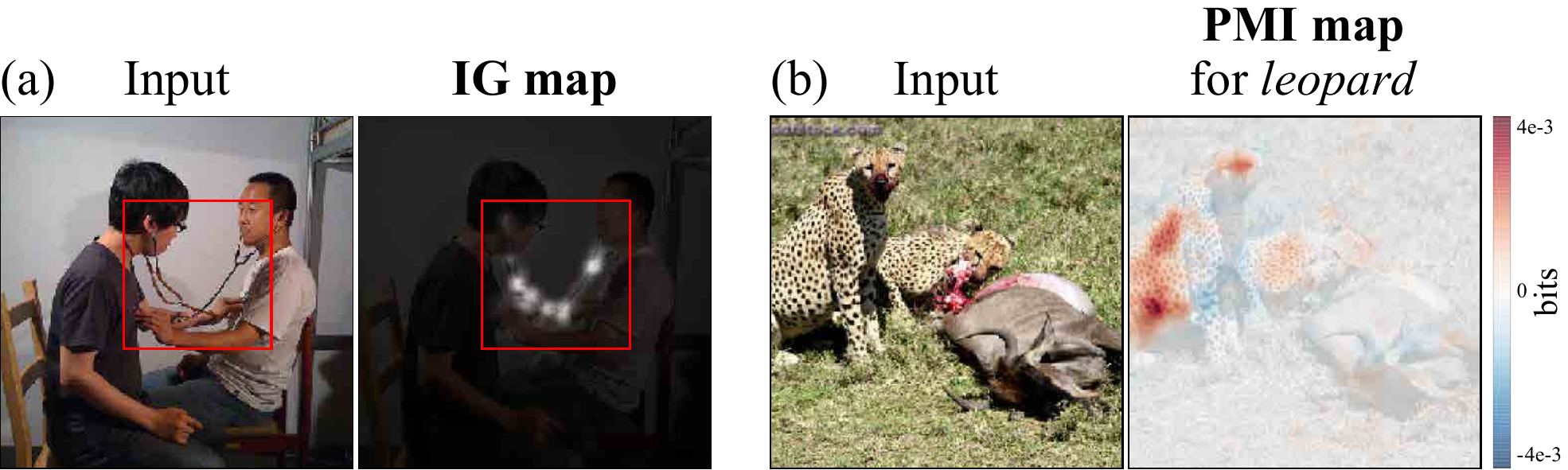}
  \end{center}
  \vspace{-5pt}
  \caption{\textbf{IG map (a) highlights the salient area on which the classifier focused. The attributions in PMI and IG maps are measured in bits (b).}}
  \label{fig:igmap_and_unit}
\end{figure}

\begin{figure}[]
  \begin{center}
    \includegraphics[width=\linewidth]{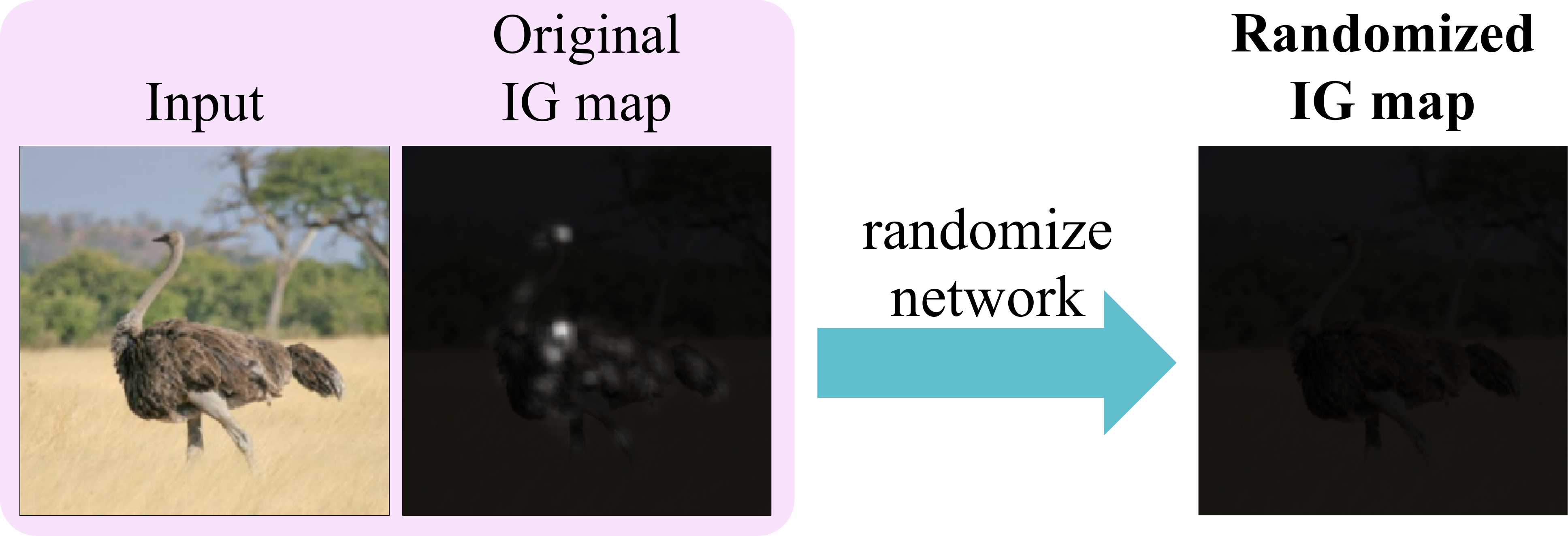}
  \end{center}
  \vspace{-15pt}
  \caption{\textbf{Sanity Checks~\cite{adebayo2018sanity}.}
   The parameters of VGG19~\cite{simonyan2014very} are successively randomized from logits to conv1 layer.
   In the end, the IG map indicates that no pixel delivers any information to the classifier.
   }\label{figure:sanity}
   \vspace{-5pt}
\end{figure}

\setlength{\columnsep}{8pt}%
\begin{wrapfigure}{r}{0.5\linewidth}
  \vspace{-5pt}
  \begin{center}
    \includegraphics[width=\linewidth]{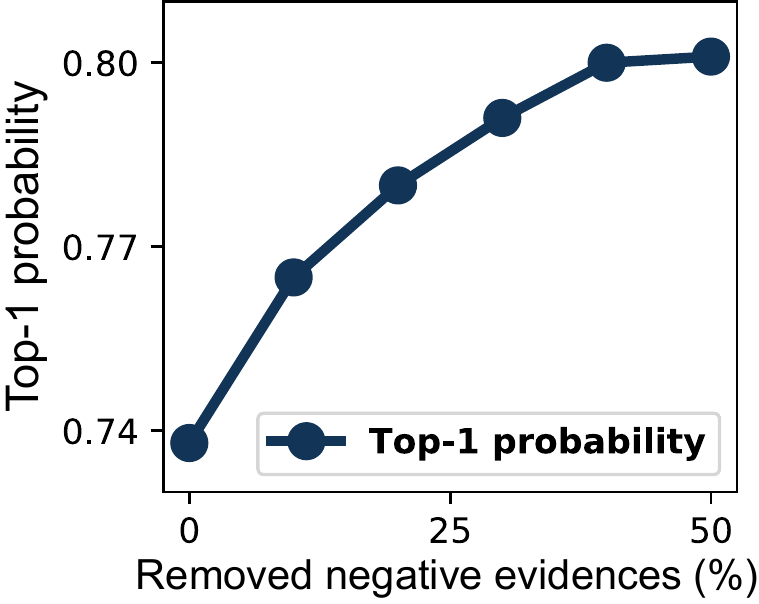}
  \end{center}
  \vspace{-18pt}
  \caption{\textbf{Removing negative evidences makes the prediction more confident.}}
  \label{fig:negative_remove}
  \vspace{-10pt}
\end{wrapfigure}
Fig.~\ref{fig:negative_remove} shows the average top-1 predicted probabilities for ImageNet test images as portions of negative evidences are gradually removed.
The result shows that the classifier gains confidence as the negative evidence is removed.
This implies that the negative evidence provided by our method is indeed negative.

\subsubsection{Class-independent explanation} \label{section:useful_igmap}

Almost all existing methods~\cite{chang2018explaining,dabkowski2017real,fong2019understanding,sundararajan2017axiomatic,zeiler2014visualizing,zintgraf2017visualizing} only provide class-specific explanations.
However, when multiple classes are probable, it is difficult to select an appropriate class-specific map to interpret the prediction.
In Fig.~\ref{fig:igmap_and_unit}(a), VGG19~\cite{simonyan2014very} predicted the image as a \textit{bassoon} ($p$ = 0.08), a \textit{stethoscope} ($p$ = 0.07), and a \textit{horn} ($p$ = 0.07) and failed to identify a dominant class.
The IG map implies that the classifier focused on the stethoscope.
Unlike in the other methods, the IG map can provide overall explanations about the region based on which the classifier made the decision.

\subsubsection{Implication of an attribution} \label{section:implication}
Mask-based methods~\cite{chang2018explaining,dabkowski2017real,fong2019understanding,fong2017interpretable} propose \textit{how to calculate} their own attribution, without a clear definition of \textit{what} the attribution is.
They quantify the relative importance of input features, but the value itself has no clear meaning.
In contrast, the PMI and IG maps measure the actual amount of \textit{Shannon's information} delivered in \textit{bits} by each feature (Fig.~\ref{fig:igmap_and_unit}(b)).
Therefore, our method provides a theoretically meaningful, yet intuitive, explanation.

\subsection{Additional remarks}
\subsubsection{Sanity Checks} \label{section:sanity}
Adebayo~\etal~\cite{adebayo2018sanity} proposed two Sanity Checks that any valid attribution method should pass.
They test if an attribution map is sensitive to both training data and the classifier parameters.
The IG map showed a clear sensitivity to the classifier parameters, as in Fig.~\ref{figure:sanity}, and both PMI and IG maps passed the two Sanity Checks.
More results are provided in the Appendix.


\begin{figure}[]
  \centering
  \includegraphics[width=\linewidth]{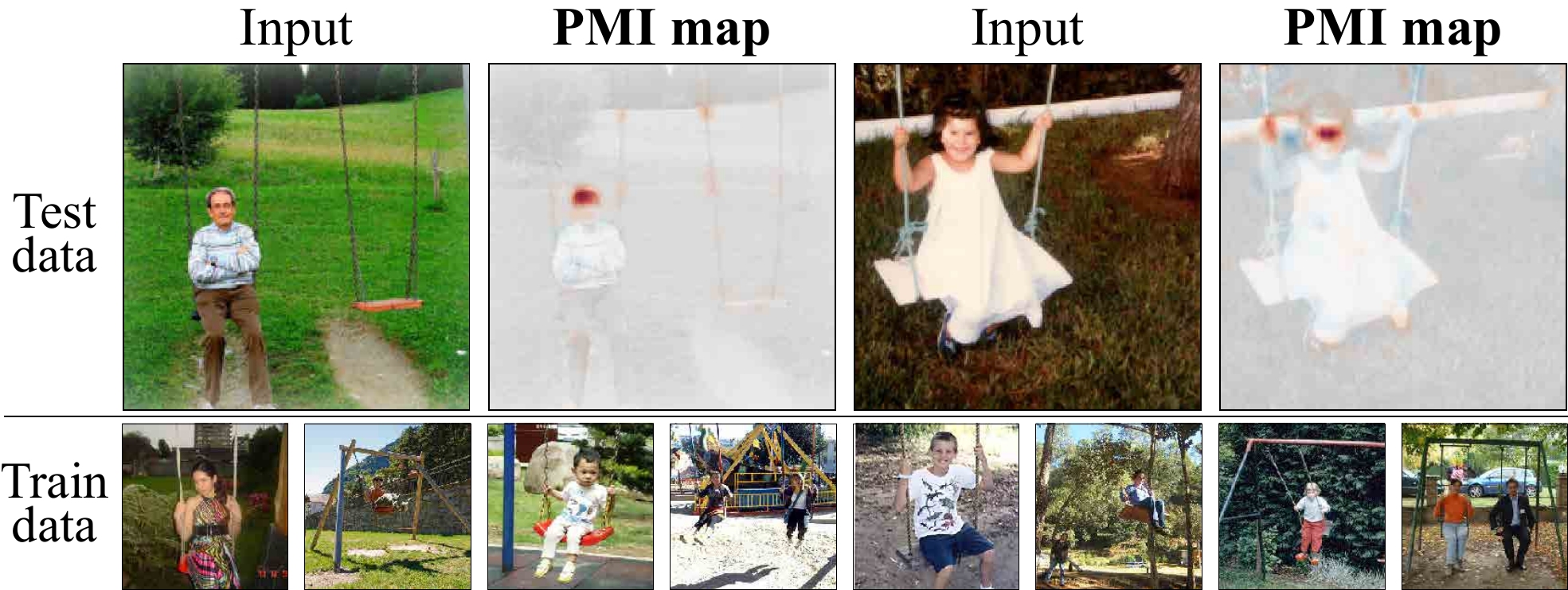}
  \vspace{-10pt}
  \caption{\textbf{VGG19~\cite{simonyan2014very} responds to a person.}
  The majority of the training data for the \textit{swing} class have a person riding on it.
  Such bias makes the classifier regard a person as a supporting evidence of the class.
  }
  \label{figure:clever_hans}
  \vspace*{\floatsep}
  
  \includegraphics[width=\linewidth]{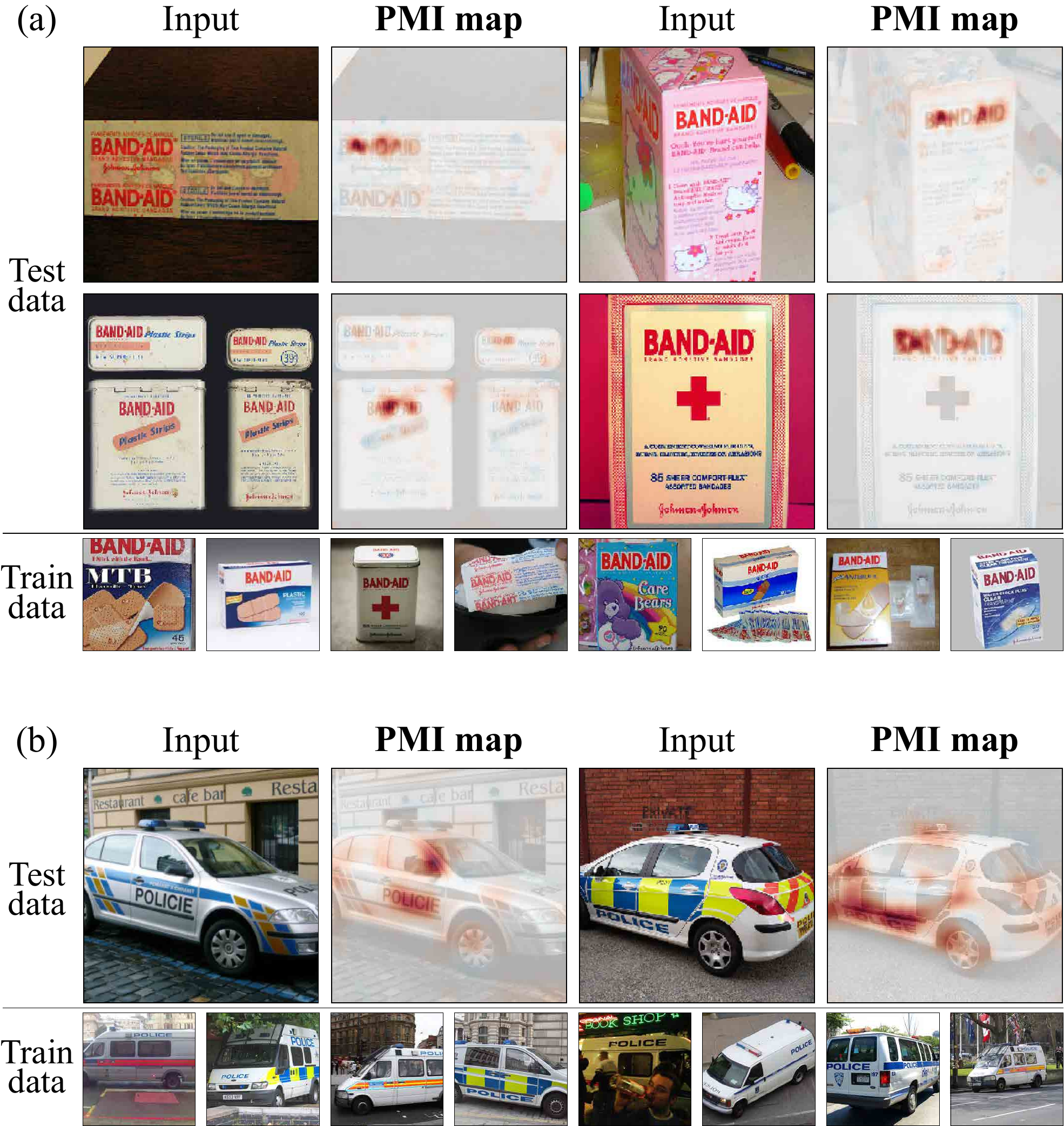}
  \vspace{-12pt}
  \caption{\textbf{VGG19~\cite{simonyan2014very} responds to texts.}
  The PMI maps in (a) suggest that the images are classified as \textit{Band Aid} class because of the ``BAND-AID" texts in the images.
  Another examples in (b) are images classified as \textit{Police van} because of ``POLICE" texts. The training images of the respective classes include the corresponding characteristic texts.}
  \label{fig:bandaid}
  \vspace{-10pt}
  
\end{figure}

\subsubsection{Biased data} \label{section:clever_hans}
In Fig.~\ref{figure:clever_hans}, the PMI maps report that the person riding a swing is more supporting feature of the class \textit{swing} than the swing itself.
This is because the majority of the training data for the class \textit{swing} contains a swing with a person riding it, as  shown in Fig.~\ref{figure:clever_hans}.
Thus, the classifier learned a strong correlation between the target class and person.
For better generalization and to reduce unwanted bias towards data, examining a trained classifier using PMI maps will be beneficial.

\subsubsection{Text cues}
In Fig.~\ref{fig:bandaid}(a), VGG19~\cite{simonyan2014very} classified the test images as the \textit{Band Aid} class, and the corresponding PMI maps suggest that the predictions are based on the red ``BAND-AID" texts in the images.
This observation can be attributed to the frequent occurrence of the similar texts in the training data.
Police van images in Fig.~\ref{fig:bandaid}(b) is another example.
\setlength{\columnsep}{8pt}%
\begin{wrapfigure}{r}{0.5\linewidth}
  \vspace{-10pt}
  \begin{center}
    \includegraphics[width=\linewidth]{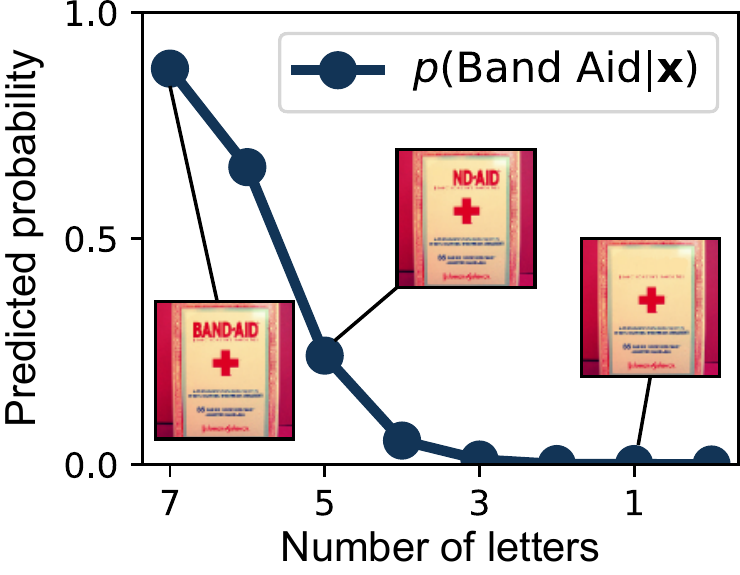}
  \end{center}
  \vspace{-18pt}
  \caption{Successive removal of letters in ``BAND-AID" text drops the predicted probability of the \textit{Band Aid} class.}
  \label{fig:bandaid_remove}
  \vspace{-10pt}
\end{wrapfigure}
To verify if the text functions as a critical cue for classification, we observe the predicted probability as the letters in the text are removed one-by-one.
Fig.~\ref{fig:bandaid_remove} shows that erasing a couple of letters yield a significant drop in the predicted probability.
This signifies that the classifier indeed responds to the text in the image.
However, the evidence is still limited to support that the classifier can \textit{read}. Rather, the classifier seems to have learned the low-level pattern as those texts are visually similar.

\begin{figure}[]
  \centering
  \includegraphics[width=\linewidth]{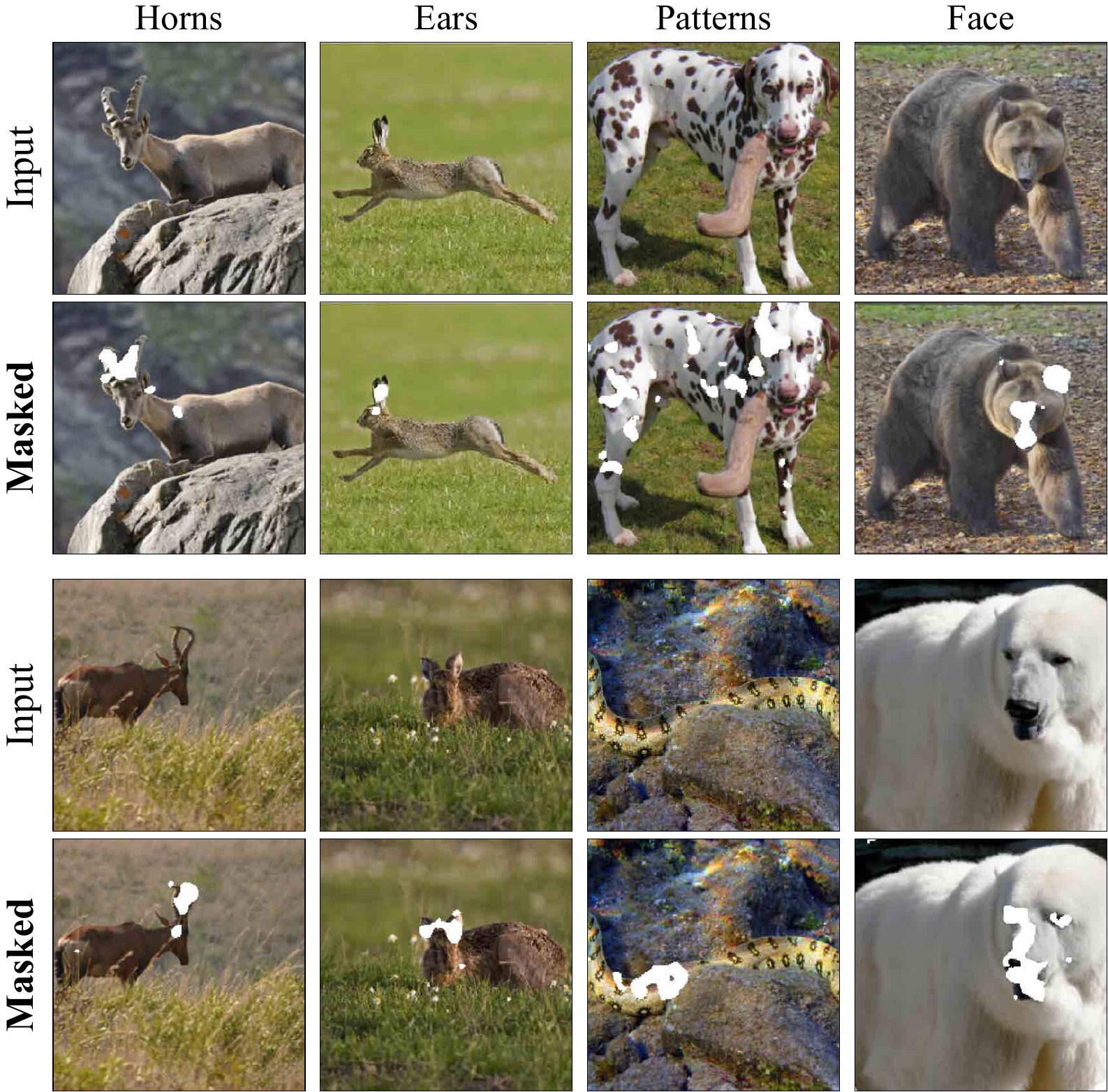}
  \vspace{-12pt}
  \caption{\textbf{VGG19~\cite{simonyan2014very} captures unique appearances.}
  The white-masked regions in the images are those with the PMI attributions larger than 20\% of the maximum value in each image.
   The masks covers distinctive animal features such as horns (\textit{ibex} and \textit{impala}), ears (\textit{hare}), and body patterns (\textit{dalmatian} and \textit{sea snake}).}
  \label{figure:animals}
  \vspace{-15pt}
\end{figure}

\subsubsection{Classifying animals} \label{section:animals}
ImageNet has a number of animal classes, and a well-trained classifier should capture the unique appearance of each animal class.
In Fig.~\ref{figure:animals}, we provided PMI maps of VGG19~\cite{simonyan2014very} for several animal classes.
The classifier successfully captured some distinctive features.
For many animals, the facial areas are particularly discriminative than the other body parts.

\subsubsection{Explaining different classifiers}
\begin{figure}[]
  \centering
  \includegraphics[width=\linewidth]{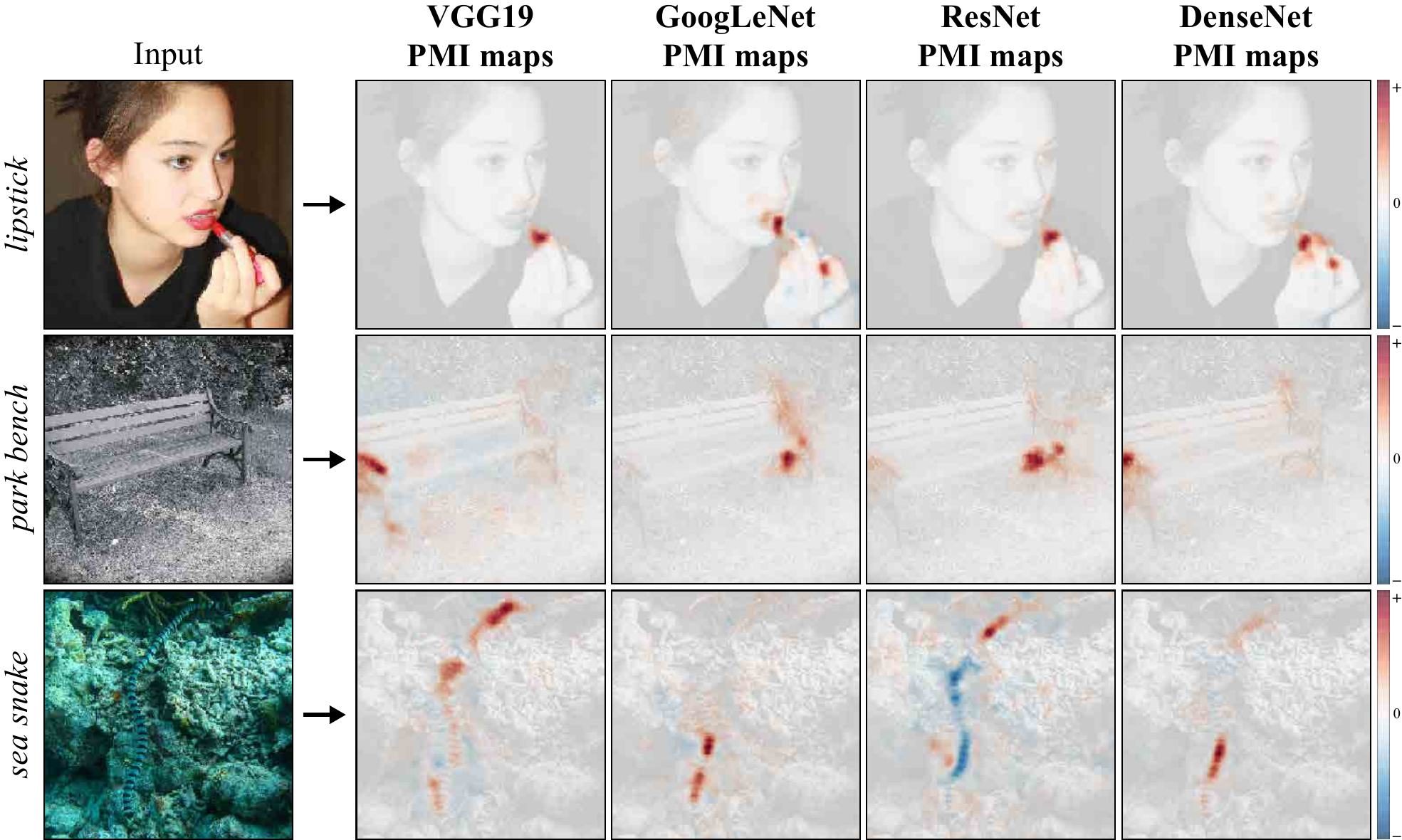}
  \vspace{-12pt}
  \caption{\textbf{Explaining different classifiers: VGG19~\cite{simonyan2014very}, GoogLeNet~\cite{szegedy2015going}, ResNet50~\cite{he2016deep}, and DenseNet121~\cite{densenet}.}
  }\label{fig:various_models}
\end{figure}

Since our method is model-agnostic, it can explain a classifier of any kind including non-differentiable one.
The PMI maps for four popular image classifiers are presented in Fig.~\ref{fig:various_models}.
The four classifiers made the same predictions for each image (provided at the left side of the input images), and the PMI maps are generated for those classes.
For some images, the classifiers focused on the similar cues (1st row), while for the other images, they made decisions based upon different areas (2nd and 3rd rows).

\subsubsection{Adversarial robustness} \label{section:adversarial_robustness}
Ghorbani \etal~\cite{ghorbani2019interpretation} proposed two attack methods (iterative and random perturbation attacks) for feature-based interpretations.
Iterative attack performs a gradient descent with respect to the input image in the direction that maximally changes the explanation, and the authors manipulated the explanation significantly while leaving the input image almost unchanged.
However, gradient signals are unattainable with our method because the sampling node followed by PatchSampler blocks backpropagation.
Only the random attack, which the authors reported to have weaker attack performance, is feasible, and hence our method is more robust against adversarial attacks than backprop-based methods such as Saliency map~\cite{simonyan2013deep} and its variants~\cite{shrikumar2017learning,sundararajan2017axiomatic}.

\section{Conclusion}
\subsection{Contributions}
In this study, we proposed a new visual explanation method based on information theory, and showed its five advantages over the existing methods.
We developed a novel approach in analyzing the input and its label as two random variables, thereby suggesting the use of theory-backed attribution methods, namely PMI and IG maps.
The improved marginalization and the use of theory-backed attribution calculation schemes provided easily interpretable and strongly convincing attribution maps.

Ribeiro \etal~\cite{ribeiro2016should} reported a trade-off between the interpretability and faithfulness of an explanation.
The most faithful explanation (\ie, the parameters themselves) lacks interpretability, and a more understandable explanation inevitably simplifies the classifier behavior, thus losing its faithfulness.
Hence, refining an attribution map by averaging out the noise \cite{smilkov2017smoothgrad} and forcing attributions to cluster or suppress artifacts using a regularizer~\cite{chang2018explaining,fong2017interpretable} gains human-interpretability in exchange for faithfulness.
By contrast, we performed no heuristic refinement for the visual appearance, yet giving easily interpretable results.

\subsection{Expandability}
Although PMI and IG maps are defined and provided for image data herein, the notions of marginalizing the input feature and measuring PMI and IG are applicable to other domains such as language modeling~\cite{devlin2019bert} and tabular data~\cite{hwang2019hexagan} using appropriate generative models.

\clearpage
\appendices
\section{}
\subsection{Pseudo-code}

  \begin{minipage}{\textwidth}
    \begin{algorithm}[H]
    \caption{Information-Theoretic Pixel Attribution}
      \label{algo:pseudo}
      \begin{algorithmic}[1]
      \State \textbf{Input} image $\xvec$, classifier $\theta$, PatchSampler $\phi$, and class $y_c$
      \State $\pmimap, \igmap \leftarrow \rm{zeros}(\rm{H}, \rm{W})$ \Comment{Initialize \textbf{PMI map} and \textbf{IG map}}
      \State Calculate $p_{\theta}(y_c|\xvec)$ \Comment{Feed-forward the classifier}
      \For{patch $\xii$ in $\xvec$}
      \ \ \ \ \State $\xhmi \leftarrow \textrm{Neighborhood}(\xii)$ \Comment{$3K \times 3K$ surrounding patch}
      \ \ \ \ \State Sample $\xti \sim p_\phi(\Xii|\xhmi)$ $N$ times \Comment{Sample from PatchSampler}
      \ \ \ \ \State Calculate $p_{\theta}(y_c|\xmi)$ using Eq. (3) \Comment{Perform marginalization}
      \ \ \ \ \State Calculate PMI and IG using Eq. (4) and Eq. (5) \Comment{Calculate attribution}
      \ \ \ \ \State Distribute PMI and IG to each pixel of $\xii$ in $\pmimap$ and $\igmap$
      \EndFor
      \State \textbf{return} $\pmimap$ and $\igmap$
      \end{algorithmic}
    \end{algorithm}                                                
  \end{minipage}

\subsection{Effect of patch size}
\begin{minipage}{\textwidth}
\begin{figure}[H]
    \centering
    \includegraphics[width=0.85\linewidth]{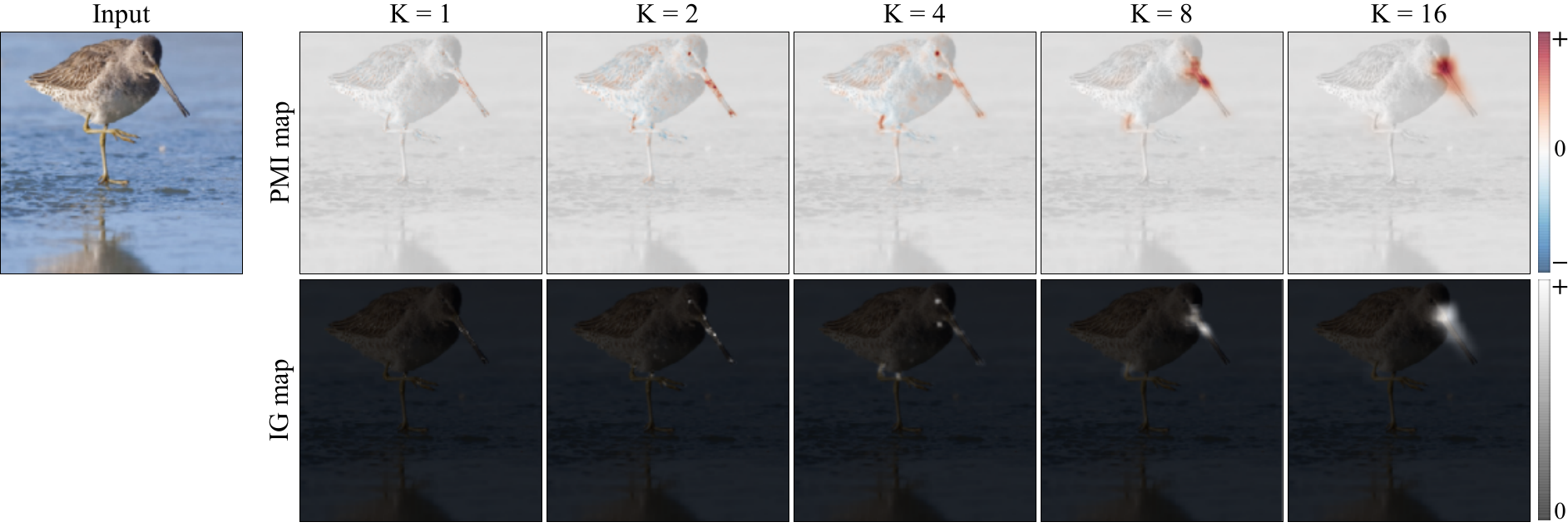}
    \caption{\textbf{Attribution maps with varying patch size, $K$.}
    VGG19~\cite{simonyan2014very} classified the image as a \textit{dowitcher} with $p=0.999$.
    As $K$ increases, our method generates more interpretable attribution maps.
    By examining these maps with proper $K$, one can easily understand that the model made a decision upon the bird's beak and facial region.
    }
    \label{fig:Various_K}
\end{figure}
\end{minipage}

\subsection{Effect of MC sample number}
\begin{minipage}{\textwidth}
\begin{figure}[H]
    \centering
    \includegraphics[width=0.85\linewidth]{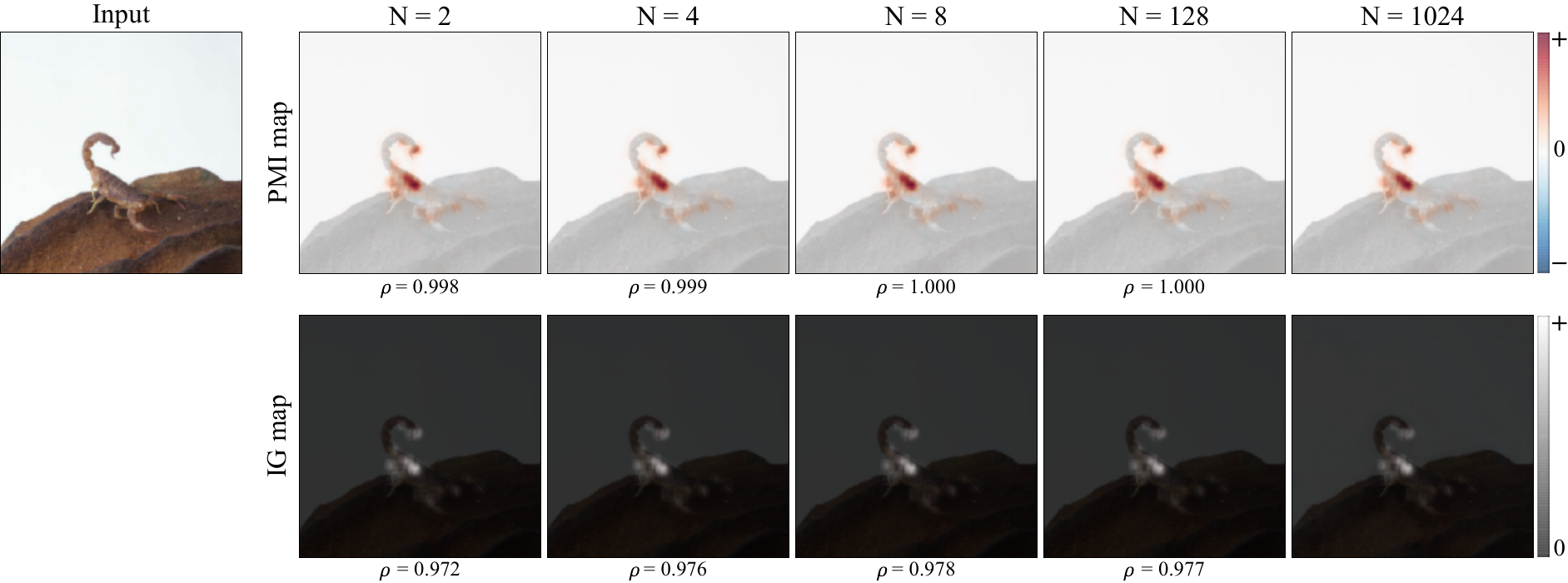}
    \caption{\textbf{Attribution maps with varying MC sample number, $N$.}
    Attribution maps when $N \leq 128$ are visually indistinguishable from maps generated using $N=1024$.
    They are almost identical in terms of the Pearson correlation coefficient ($\rho$) as well.
    Accordingly, we chose $N=8$ as a point of compromise between computational efficiency and approximation accuracy and used this value throughout the study.
    }
    \label{fig:Various_N}
\end{figure}
\end{minipage}
\clearpage

\subsection{Sanity Checks}
\begin{turn}{90}
\noindent
\begin{minipage}{0.95\textheight}
\begin{figure}[H]
\vspace{-25pt}
\centering
\includegraphics[width=0.85\linewidth]{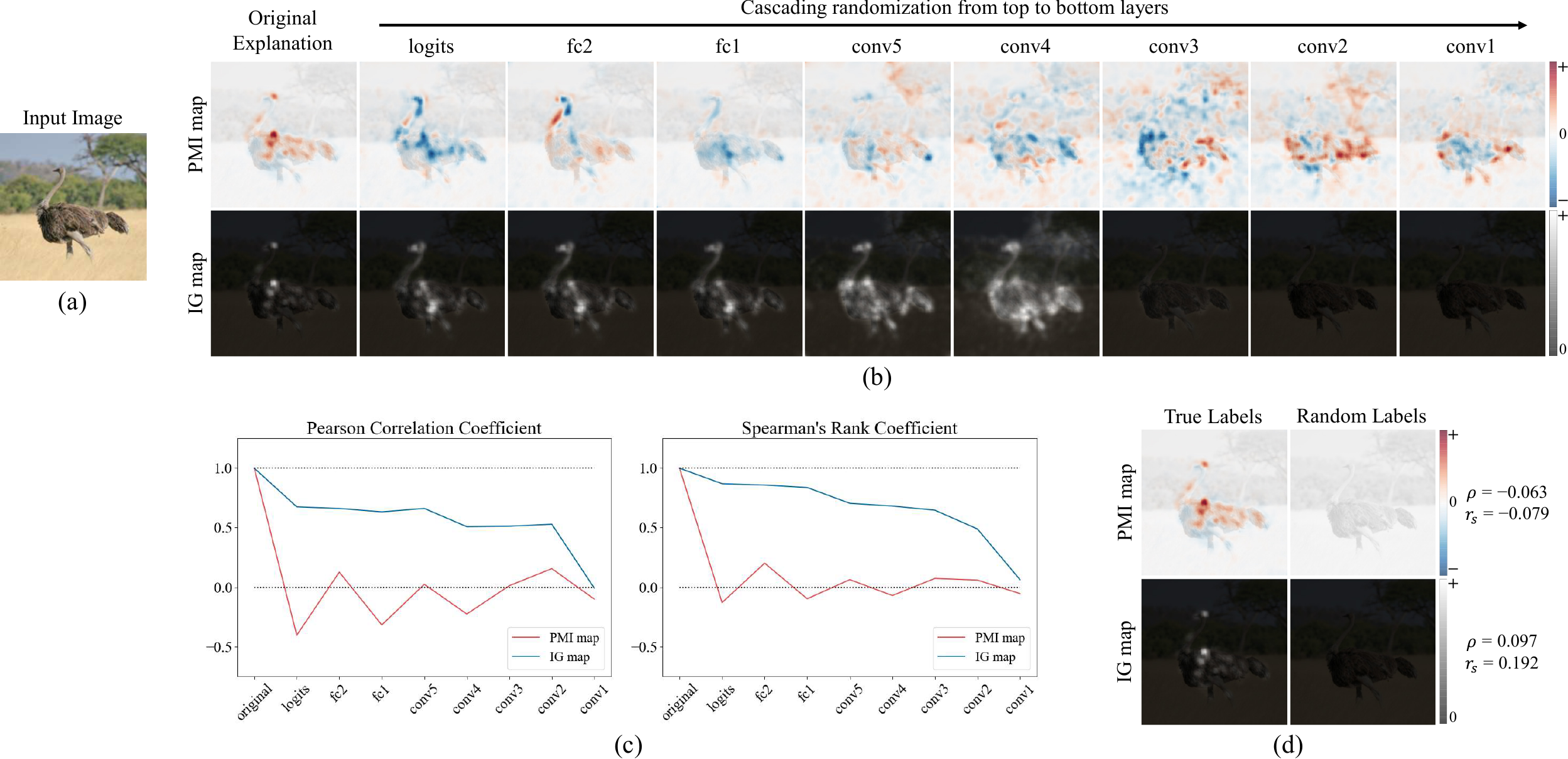}
\caption{
\textbf{An input image (a), results from the parameter randomization test~\cite{adebayo2018sanity} (b, c), and the label randomization test~\cite{adebayo2018sanity} (d).}
In a parameter randomization test, weights of VGG19~\cite{simonyan2014very} are randomized from top to bottom layers.
PMI and IG maps for each randomized classifier are depicted in (b).
As each layer is successively randomized, both attribution maps become dissimilar to the original explanation.
Pearson correlation coefficient and Spearman's rank coefficient between each map and the original maps are plotted in (c).
PMI and IG maps for a fully randomized classifier are completely different from the original maps; both coefficients are close to zero.
Thus they passed the first test.
In a label randomization test, VGG19 is retrained to fit the ImageNet training images with random labels.
Both attribution maps for the deformed classifier look completely different from the original maps. 
They imply that no pixels provided useful information \textit{overall} nor supported the \textit{ostrich} class.
Indeed, numerical comparison using Pearson correlation coefficient ($\rho$) and Spearman's rank coefficient ($r_s$) confirms this point.
Therefore, PMI and IG maps passed both Sanity Checks~\cite{adebayo2018sanity}.
}
\label{fig:rand_param}
\end{figure}
\end{minipage}
\end{turn}
\clearpage
\section{Implementation Details}
\subsection{Environments}
The proposed method is implemented using TensorFlow~\cite{abadi2016tensorflow} version 1.12, on a machine consisting of an Intel i7-6850K CPU and a GeForce GTX 1080 Ti GPU.

\subsection{Running time}
The generation of a pair of PMI and IG maps of VGG19~\cite{simonyan2014very} for one input image requires less than thirty minutes using a single GPU.
Note that the running time is linear in $N$, and reducing $N$ to a quarter of the used value (2) gives almost identical attribution maps.
Moreover, additional approximations, such as increasing the patch stride, can further reduce the computation time to a couple of minutes.

\subsection{PatchSampler}
PatchSampler predicts pixel values of a patch from a larger patch surrounding it.
We model PatchSampler as a stack of the convolutional (conv.) layers as follows: conv9-256, conv9-256, maxpool2, crop out 2 pixels from each edge, conv8-256, conv8-256, conv8-256, conv8-256, conv8-256, and conv8-768.
Every conv. and maxpool operations used stride size of one (conv9-256 means a conv. layer with kernel size 9 yielding 256 feature maps).
The channels in the last layer are grouped into three, representing RGB channel for 256 possible values.
Therefore, the softmax activations are applied for each group of channels.
The output of PatchSampler is three groups of 256-class categorical distributions for each pixel and each RGB channel.
PatchSampler is then trained to predict the pixel values using an ImageNet training set and multi-class classification loss.
All convolutional layers except for the last layer are followed by a LeakyReLU~\cite{maas2013rectifier} activation with $\alpha=0.2$.

\subsection{Classifiers}
The parameter weights of the analyzed classifiers were downloaded from PyTorch model zoo \footnote{\url{https://pytorch.org/docs/stable/model_zoo.html\#module-torch.utils.model_zoo}}.

\subsection{Methods}
The implementations of other attribution maps are downloaded from various online sources as follows.
Prediction Difference Analysis (PDA)~\cite{zintgraf2017visualizing}: \url{https://github.com/lmzintgraf/DeepVis-PredDiff};
Saliency map (Gradients)~\cite{simonyan2013deep}, Integrated Gradients~\cite{sundararajan2017axiomatic}, and $\epsilon$-LRP~\cite{bach2015pixel}: \url{https://github.com/marcoancona/DeepExplain}; Grad-CAM~\cite{selvaraju2017grad}: \url{https://github.com/jacobgil/pytorch-grad-cam};
Real Time Image Saliency~\cite{dabkowski2017real}: \url{https://github.com/PiotrDabkowski/pytorch-saliency}
; Meaningful Perturbation~\cite{fong2017interpretable}: \url{https://github.com/jacobgil/pytorch-explain-black-box}
; FIDO~\cite{chang2018explaining}: \url{https://github.com/zzzace2000/FIDO-saliency}
; Extremal Pertubation~\cite{fong2019understanding}: \url{https://github.com/facebookresearch/TorchRay}
; RISE~\cite{petsiuk2018rise}: \url{https://github.com/eclique/RISE}.
Occlusion~\cite{zeiler2014visualizing} was implemented by us.
SSR loss was used for a generation of FIDO maps because the loss is reported to be less susceptible to artifacts than SDR loss~\cite{chang2018explaining}.

\section{Additional Remarks}
\subsection{Backprop-based method}
Since Simonyan \etal~\cite{simonyan2013deep} first suggested Saliency map to visualize attribution by using a gradient of class score with respect to the input, successive works have improved its visual quality by reducing the noise with averaging~\cite{smilkov2017smoothgrad} and by using integration~\cite{sundararajan2017axiomatic}.
In addition, some works have changed the back-propagation rule by using various heuristic methods such as Deconv~\cite{zeiler2014visualizing}, Layer-wise Relevance Propagation (LRP)~\cite{bach2015pixel}, and Guided Back Propagation~\cite{springenberg2014striving}, thereby producing visually appealing results.
Because these methods calculate an attribution map through several backpropagations, they are called backprop-based methods.

As indicated by Dabkowski~\etal~\cite{dabkowski2017real}, the backprop-based methods are fast enough to be real-time, but their quality is limited.
Moreover, some of these methods fail Sanity Checks~\cite{adebayo2018sanity} by showing invariance to classifier parameters and to the training data.
Therefore, the failed methods are proved to be inadequate in many domains.
Moreover, Nie~\etal~\cite{nie2018theoretical} theoretically showed that Deconv~\cite{zeiler2014visualizing} and Guided Back Propagation~\cite{springenberg2014striving} simply generated a map similar to the input image, rather than providing an explanation for the classifier.

\subsection{Mask-based method}
Dabkowski \etal~\cite{dabkowski2017real} attempted to find an attribution map having a different implication.
Their attribution map is a mask that covers a relevant region of the target object.
They generated the mask in real-time by training a mask-generating model in advance.
In another approach \cite{fong2017interpretable,chang2018explaining,fong2019understanding}, similar masks were optimized iteratively using a gradient descent, in which additional loss terms were used to force the mask to cluster together and suppress high-frequency components.
Such masks tend to cover the entire object in the image rather than only the most decisive portion within the object.
For example, given an image of a \textit{car}, mask-based methods cover the entire car area, whereas in other approaches \cite{zeiler2014visualizing,zintgraf2017visualizing}, a high score will be assigned only to the most salient subregions, such as the wheels.

\subsection{Shapley-based method}
Shapley value~\cite{shapley201617} quantifies the contribution of each player in a coalition game.
Following this game-theoretic approaches, Lundberg~\etal~\cite{lundberg2017unified} adopted Shapley value to estimate the feature importance.
They estimate the expected amount of prediction change if each feature is masked out under all possible context.
Frye~\etal~\cite{frye2020shapley} takes the similar approach while using a generative model to keep the masked data on the data manifold.
The use of a generative model by the authors has a similar motivation to that of PatchSampler in the perturbation-based methods.
However, Kumar~\etal~\cite{kumar2020problems} claimed the inappropriateness of Shapley value as an attribution method.

\ifCLASSOPTIONcaptionsoff
  \newpage
\fi

\bibliographystyle{IEEEtran}
\bibliography{references}

\begin{IEEEbiography}[{\includegraphics[width=1in,height=1.25in,clip,keepaspectratio]{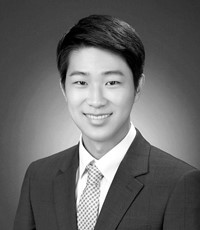}}]{Jihun Yi}
received the B.S. degree in electrical and computer engineering from Seoul National University, Seoul, South Korea, in 2017, where he is currently pursuing the Ph.D. degree in electrical and computer engineering. His research interests include deep learning, anomaly detection, and explaninable AI.
\end{IEEEbiography}

\begin{IEEEbiography}[{\includegraphics[width=1in,height=1.25in,clip,keepaspectratio]{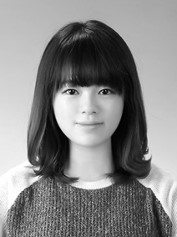}}]{Eunji Kim}
received the B.S. degree in electrical and computer engineering from Seoul National University, Seoul, South Korea, in 2018, where she is currently pursuing the integrated M.S./Ph.D. degree in electrical  and  computer  engineering. Her research interests include artificial intelligence, deep learning, and computer vision.
\end{IEEEbiography}

\begin{IEEEbiography}[{\includegraphics[width=1in,height=1.25in,clip,keepaspectratio]{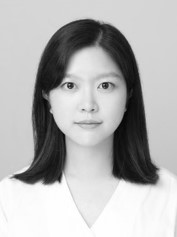}}]{Siwon Kim}
received her B.S. degrees in electrical and computer engineering from Seoul National University in Seoul, Korea in 2018. Currently, she is pursuing an integrated M.S./Ph.D. degree  in  electrical  and  computer  engineering  at Seoul National University. Her research interests include artificial intelligence, deep learning and biomedical applications.
\end{IEEEbiography}

\begin{IEEEbiography}[{\includegraphics[width=1in,height=1.25in,clip,keepaspectratio]{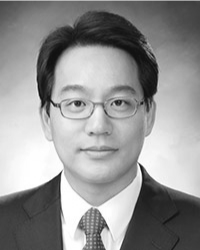}}]{Sungroh Yoon}
(S’99–M’06–SM’11)
received the B.S. degree in electrical engineering
from Seoul National University, South Korea,
in 1996, and the M.S. and Ph.D. degrees in
electrical engineering from Stanford University,
CA, USA, in 2002 and 2006, respectively. From
2016 to 2017, he was a Visiting Scholar with
the Department of Neurology and Neurological
Sciences, Stanford University. He held research
positions at Stanford University and Synopsys,
Inc., Mountain View, CA, USA. From 2006 to 2007, he was with Intel
Corporation, Santa Clara, CA, USA. He was an Assistant Professor with
the School of Electrical Engineering, Korea University, from 2007 to 2012.
He is currently a Professor with the Department of Electrical and Computer Engineering, Seoul National University. His current research interests include machine learning and artificial intelligence. He was a recipient of
the SNU Education Award, in 2018, the IBM Faculty Award, in 2018,
the Korean Government Researcher of the Month Award, in 2018, the BRIC
Best Research of the Year, in 2018, the IMIA Best Paper Award, in 2017,
the Microsoft Collaborative Research Grant, in 2017, the SBS Foundation
Award, in 2016, the IEEE Young IT Engineer Award, in 2013, and many
other prestigious awards.
\end{IEEEbiography}

\end{document}